\documentclass[
]{ceurart}

\usepackage{multicol}
\usepackage{multirow}
\usepackage{listings}
\begin{document}

\copyrightyear{2023}
\copyrightclause{Copyright for this paper by its authors.
  Use permitted under Creative Commons License Attribution 4.0
  International (CC BY 4.0).}

\conference{In A. Martin, K. Hinkelmann, H.-G. Fill, A. Gerber, D. Lenat, R. Stolle, F. van Harmelen (Eds.), 
Proceedings of the AAAI 2023 Spring Symposium on Challenges Requiring the Combination of Machine Learning and Knowledge Engineering (AAAI-MAKE 2023), Hyatt Regency, San Francisco Airport, California, USA, March 27-29, 2023.}

\title{Fine-Grained ImageNet Classification in the Wild}

\author[1]{Maria Lymperaiou}[%
orcid=0000-0001-9442-4186,
email=marialymp@islab.ntua.gr]
\author[1]{Konstantinos Thomas}[%
orcid=0000-0002-7489-7776,
email=kthomas@islab.ntua.gr]
\address[1]{AILS Lab,
School of Electrical and Computer Engineering,
National Technical University of Athens}

\author[1]{Giorgos Stamou}[orcid= 0000-0003-1210-9874, email=gstam@cs.ntua.gr]

\begin{abstract}
Image classification has been one of the most popular tasks in Deep Learning, seeing an abundance of impressive implementations each year. However, there is a lot of criticism tied to promoting complex architectures that continuously push performance metrics higher and higher. Robustness tests can uncover several vulnerabilities and biases which go unnoticed during the typical model evaluation stage. So far, model robustness under distribution shifts has mainly been examined within carefully curated datasets. Nevertheless, such approaches do not test the real response of classifiers in the wild, e.g. when uncurated web-crawled image data of corresponding classes are provided. In our work, we perform fine-grained classification on closely related categories, which are identified with the help of hierarchical knowledge. Extensive experimentation on a variety of convolutional and transformer-based architectures reveals model robustness in this novel setting. Finally, hierarchical knowledge is again employed to evaluate and explain misclassifications, providing an information-rich evaluation scheme adaptable to any classifier.
\end{abstract}


\maketitle

\section{Introduction}
ImageNet \cite{imagenet} has been one of the most popular image classification datasets in literature, inspiring a variety of popular model implementations for over a decade. The first significant breakthrough in ImageNet classification was marked with AlexNet \cite{alexnet}, a convolutional neural network (CNN) for image classification that greatly outperformed its competitors. Ever since various CNN-based implementations continued pushing accuracy scores even higher \cite{image-classification-survey}. 

The local nature of convolutional filters that cannot capture long-range visual dependencies was suspected to hinder further improvements in performance, demanding the exploration of alternative architectural choices. To this end, attention mechanisms that have successfully served Natural Language Processing \cite{transformer} appear as a promising substitute to convolutions, as they are able to detect spatially distant concepts and assign appropriate importance weights to them. Indeed, the adaptation of the Transformer \cite{transformer} for visual tasks, led to the introduction of the Visual Transformer (ViT) \cite{vit}, which divides the image into visual patches and processes them similarly to how the original Transformer handles words.
Consequently, transformer-based image classifiers emerged
\cite{coca, modelsoups, swin2, beit}, reaching unprecedented state-of-the-art results.

Even though so much effort is invested to perpetually improve model performance by employing more and more refined architectures and techniques, inevitably increasing the demand for computational resources necessary for training, there are still some open questions regarding the ability of such models to properly handle distribution shifts. Distribution shifts refer to testing an already trained model on a data distribution that diverges from the one the model was trained on. The analysis of distribution shifts has gained interest in recent years \cite{shift1, robustness5, shift2, shift3, robustness4}, as a crucial step towards enhancing model robustness. Most of these endeavors apply pixel-level perturbations to artificially influence the distribution under investigation. 
Nevertheless, the highly constrained setting of artificial distribution shifts excludes various real-world scenarios, impeding robust generalization of image classifiers. In this case, natural shifts \cite{robustness1, robustness3, robustness7, robustness8} are more representative. They usually require the creation of a \textit{curated} dataset containing image variations such as changes in viewpoint or object background, rotations, and other minor changes. Both synthetic and natural shifts can comprise data augmentation techniques, which aid the development of robust models when incorporated during training \cite{train0,train1,train4, train2}.

So far, there is no approach testing image classification 'in the wild', where \textit{uncurated} images corresponding to pre-defined dataset labels are encountered. We argue that this is a real-world user-oriented scenario, where totally new images corresponding to ImageNet labels need to be appropriately classified. For example, an image of a cat found on the web may significantly differ from ImageNet cat instances, even when popular distribution shifts are taken into account. Even though a human can identify a cat present in an image with satisfactory confidence, we question whether an image classifier can do so; the unrestricted space of possible variations of uncurated images demands advanced generalization capabilities to properly understand the real discriminative characteristics of an ImageNet class without getting distracted from extraneous features.

The problem of classification 'in the wild' becomes even more difficult when fine-grained classification needs to be performed, as distinguishing between closely related categories relies on detailed discriminative characteristics, which may be less prevalent in uncurated settings. For example, siamese and persian cat races present many visual similarities, increasing the potential risk of learning and reproducing dataset biases, especially when distribution shifts are present. We can attribute this risk to the fact that existing classifiers lack \textit{external} or \textit{domain} \textit{knowledge}, which can help humans discriminate between closely related categories.

To sum up, in our current paper we aspire to answer the following questions:
\begin{enumerate}
    \item How do different models, pre-trained on ImageNet or web images, behave on uncurated image sets crawled from Google images (given ImageNet labels as Google queries)? We target this question by producing a novel natural \textit{distribution shift} based on uncurated web images upon which we evaluate various image classifiers.
    \item How does hierarchical knowledge help with evaluating  classification results since several ImageNet categories are hierarchically related? We attempt to verify to which extent the assumption that \textit{the lack of external knowledge limits the generalization capabilities of classifiers} holds. Thus, we leverage WordNet \cite{wordnet} to discover neighbors of given terms and test whether classifiers struggle with discriminating between closely related classes.
    \item Can evaluation of classification be \textit{explainable}? Knowledge sources, such as WordNet can reveal the semantic relationships between concepts (ImageNet classes), providing possible paths connecting frequently confused classes.
\end{enumerate}

Our code can be found at \href{https://github.com/marialymperaiou/classification-in-the-wild}{https://github.com/marialymperaiou/classification-in-the-wild}.

\section{Related work} 

\paragraph{Image classifiers} With the outburst of neural architectures for classification tasks, Computer Vision has been one of the fields most benefited from recent developments. Convolutional classifiers (CNN) is a well-established backbone, with first successful endeavors \cite{alexnet} already paving the way for more refined architectures, 
such as VGG \cite{vgg}, Inception \cite{inception}, ResNet \cite{resnet}, Xception \cite{Xception}, InceptionResnet \cite{inceptionresnet} and others \cite{image-classification-survey}. There is some criticism around the usage of CNNs for image classification, even though some contemporary endeavors such as ConvNext \cite{convnext} revisit and insist on the classic paradigm, providing advanced performance.
The rapid advancements that the Transformer framework \cite{transformer} brought via the usage of self-attention mechanisms, widely replacing prior architectures for Natural Language Processing applications, inspired the usage of similar models for Computer Vision as an answer to the aforementioned criticism \cite{vit-survey}. Thus, Vision Transformers (ViTs) \cite{vit} built upon \cite{transformer} set a new baseline in literature; ever since, several related architectures emerged. In general, transformer-based models rely on an abundance of training data to ensure proper generalization. This requirement was relaxed in DeiT \cite{deit}, enabling learning on medium-sized datasets. Further development introduced novel transformer-based architectures, such as BeiT \cite{beit}, Swin \cite{swin} and RegNets \cite{regnet}, which realize specific refinements to boost performance. Overall, it has been proven that ViTs are more robust compared to classic CNN image classifiers \cite{robustvit}. In our work, we verify the degree this claim holds by testing CNN and transformer-based classifiers on the uncurated fine-grained setting.

\paragraph{Robustness under distribution shifts}
Generalization capabilities of existing image classifiers have been a crucial problem \cite{robustness0}, currently addressed from a few different viewpoints. Artificial corruptions \cite{robustness6, robustness4, robustness2, robustness3, robustness5} or natural shifts \cite{robustness1, nae} on \textit{curated} data have already exposed biases and architectural vulnerabilities. Adversarial robustness \cite{adversarial3, adversarial1, adversarial5, adversarial2, adversarial4} is a related field where models are tested against adversarial examples, which introduce imperceptible though influential perturbations on images. 
 Contrary to such attempts, we concentrated around naturally occurring distribution shifts stemming from \textit{uncurated} image data.
Regarding architectural choices, many studies perform robustness tests attempting to resolve the CNN vs Transformer contest \cite{robustvit, impartial, cnn-or-transformer}, while other ventures focus on interpreting and understanding model robustness \cite{understand-robust, understand-robust2, understand-robust3}. In our approach, by experimenting with both CNN and transformer-based architectures we adopt such research attempts to the \textit{uncurated} setting. 

\section{Method}
\begin{figure}[h!]
    \centering
    \includegraphics[width=0.95\textwidth]{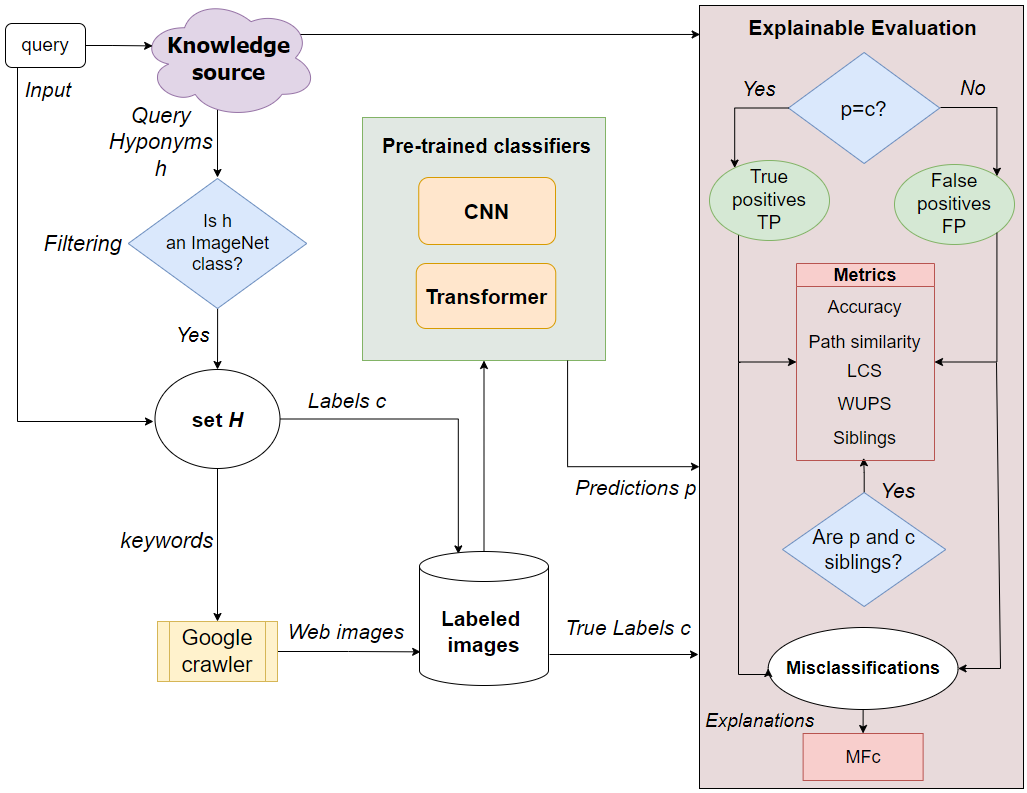}
    \caption{Outline of our method.}
    \label{fig:outline}
\end{figure}

The general workflow of our method (Figure \ref{fig:outline}) consists of three stages. First, the dataset should be constructed by gathering common terms (queries) and their subcategories which exist as ImageNet classes. Images corresponding to those terms are crawled from Google search. In the second stage, various pre-trained classifiers are utilized to classify crawled images. The hierarchical relationships between the given classes are reported to enrich the evaluation process. Finally, all semantic relationships between misclassified samples are gathered to extract explanations and quantify how much, falsely predicted classes, diverge from their ground truth.

\paragraph{Dataset creation}
\label{creation}
We start by gathering user-defined common words regarding visual concepts as \textit{queries}, which will act as starting points towards extracting subcategories. The WordNet hierarchy \cite{wordnet} is used to provide the subcategories, via the hypernym-hyponym (IsA) relationships, which refer to more general or more specific concepts respectively. For example, given the query 'car', its hypernym is 'motor vehicle' ('car' IsA 'motor vehicle'), while its hyponyms are 'limousine' ('limousine' IsA 'car'), 'sports car' ('sports car' IsA 'car') and other specific car types. Therefore, we map queries on WordNet to obtain all their immediate hyponyms, constructing a \textit{hyponyms set} $H$. We then filter out any hyponyms not belonging to ImageNet class labels.

The filtered categories of $H$ among the initial query are provided as search terms to a web crawler suitable for searching Google images. We set a predefined threshold $k$ for the number of Google images returned so that we evaluate classifiers on categories containing almost equal numbers of samples. This is necessary since some popular categories may return way more Google images compared to others. We will experiment with several values of \textit{k}, thus influencing the tradeoff between relevance to the keyword and adequate dataset size. The retrieved images comprise a labeled dataset $D$, with the keywords as labels.

\paragraph{Classification}
We consider a variety of image classifiers to test their ability for fine-grained classification on uncurated web images. We commence our experimentation with convolutional-based models as baselines, which have generally been considered to be less robust against distribution shifts and other perturbations \cite{robustvit}, and we proceed with recent transformer-based architectures. We perform no further training or fine-tuning on the selected models.

For each model, we perform inference on the crawled images that constitute our dataset, as explained in the previous paragraph. We implement a rich evaluation scheme to capture various insights of the classification process. Accuracy is useful as a benchmark metric to compare our findings with expected classification results. WordNet similarity functions offer valuable information about misclassifications; for example, let's assume that the true label of a sample is 'cat' and the classifier predicts the label 'dog' in one case and the label 'airplane' in another case. Intuitively, we hypothesize that a 'cat' is more closely related to a 'dog' than an 'airplane' since they are both animals. This human intuition is reflected in the WordNet hierarchy, thus assigning a different penalty depending on the concept relevance within the hierarchy. 

This concept-based evaluation can be realized using the following WordNet functions: path similarity, Leacock-Chodorow Similarity (LCS), and Wu-Palmer Similarity (WUPS). \textbf{Path similarity} evaluates how similar two concepts are, based on the shortest path that connects them within the WordNet hierarchy. It can provide values between 0 and 1, with 1 denoting the maximum possible similarity score. \textbf{LCS} also seeks for the shortest path between two concepts but additionally regards the depth of the taxonomy. Specifically, equation \ref{eq:lcs} mathematically describes LCS between two concepts $c_1$ and $c_2$:
\begin{equation} \label{eq:lcs}
    LCS=-\log\frac{path(c_1, c_2)}{2\cdot d}
\end{equation}
where $path(c_1, c_2)$ denotes the shortest path connecting $c_1$ and $c_2$ and $d$ refers to the taxonomy depth. Higher LCS values indicate higher similarity between concepts.
\textbf{WUPS} takes into account the depth that the two concepts $c_1$ and $c_2$ appear in WordNet taxonomy and the depth of their most specific common ancestor node, called Least Common Subsumer. Higher WUPS scores refer to more similar concepts. For each of the path similarity, LCS, and WUPS metrics we obtain an average value over the total number of samples of the constructed dataset $D$.

Moreover, we report the percentage of \textit{sibling concepts} among misclassifications. Two concepts are considered to be siblings if they share an immediate (1 hop) parent. For example, the concepts 'tabby cat' and 'egyptian cat' share the same parent node ('domestic cat').
It is highly likely that a classifier is more easily confused between two sibling classes, thus providing false positive (FP) predictions closely related to the ground truth (GT) label. Therefore, a lower number of siblings denotes reduced classification capacity compared to models of higher siblings percentage.

\paragraph{Explanations} are provided during the evaluation stage, aiming to answer \textit{why a pre-trained classifier cannot correctly classify uncurated images belonging to a class $c$.}

FP predictions contain valuable information regarding which classes are confused with the GT. The per-class misclassification frequency (MF) refers to the percentage of occurrences of each false positive class $f$ within the total number of false positive instances. Thus, given a dataset with $N$ classes, $c$ as the ground truth class and $f$ as one of the false positive classes,
the misclassification frequency for the $c\rightarrow f$ misclassification is:

\begin{equation}
    MF_c = \frac{FP_{i=f}}{\displaystyle\sum_{i=0}^{i=N}FP_i}\cdot 100\%
\end{equation}

$MF$ scores can be extracted for all $f\neq c$ FP classes so that the most influential misclassifications are discovered.
Higher $MF$ scores denote some classifier tendency to choose the FP class over the GT one, therefore indicating either a classifier bias or an annotation error in the dataset. Specifically, a classifier bias refers to consistently classifying samples from class $c$ as samples of class $f$, given that the annotation is the best possible. Of course, such a requirement cannot be always satisfied, especially when expert annotators are needed, as may happen in the case of fine-grained classification. On the other hand, since our explainable evaluation approach is able to capture such misclassification patterns, it is not necessary to attribute the source of misclassification beforehand. Human annotators can be employed at a later stage, identifying and verifying the source of misclassifications.

\section{Experiments}
In all following experiments, we selected a threshold of $T$=50 crawled images per class. We will present results on a random initial query as a proof-of-concept to demonstrate our findings. For this reason, we provide the query 'cat', which returns the following WordNet hyponyms (also corresponding to ImageNet labels): 

$H$=\{'angora cat',
 'cougar cat',
 'egyptian cat',
 'leopard cat',
 'lynx cat',
 'persian cat',
 'siamese cat',
 'tabby cat',
 'tiger cat'\}

The same experimentation can be replicated for other selected queries, as long as they can be mapped on WordNet.

\begin{table}[htp!]
\caption{Classification results using CNNs. Bold entries denote lower accuracy compared to best model accuracy. }
\label{tab:cnn-results}
\begin{tabular}{c|ccc|cccc}
\toprule
Model & Label & Accuracy$\uparrow$ & Siblings$\uparrow$ & Label  & Accuracy$\uparrow$ & Siblings$\uparrow$ \\
\midrule
ResNet50 & \multirow{8}{5em}{tabby cat}& \textbf{50.00}\% & 24.00\% & \multirow{8}{5em}{siamese cat} & 90.00\% & 0.00\%  \\
ResNet101 && \textbf{52.00}\% & \underline{41.67\%}  && 88.00\% & 16.67\% \\
ResNet152 && \textbf{50.00}\% & 12.00\%  && 90.00\% & 20.00\%\\
VGG16 && \textbf{38.00}\% & 38.71\%  && 82.00\% & 11.11\%\\
VGG19 && \textbf{50.00}\% & 32.00\%  && 88.00\% & 16.67\%\\
MobileNetV2 && \textbf{2.00}\% & 2.04\%  && \textbf{4.00}\% & 0.00\% \\
EfficientNet && \textbf{10.00}\% & 33.33\% && \underline{96.00\%} & \underline{100.00\%}\\
ConvNext && \underline{\textbf{60.00}\% }& 15.00\% && 92.00\% & 75.00\% \\
\midrule
ResNet50 &\multirow{7}{5em}{lynx cat}& 82.00\% & 0.00\% & \multirow{7}{5em}{cougar cat}& 84.00\% & 0.00\%\\
ResNet101 && 84.00\% & 0.00\% && 78.00\% & 0.00\% \\
ResNet152 && 86.00\% & 0.00\% && 88.00\% & 0.00\% \\
VGG16 && 82.00\% & 0.00\% && 86.00\% & 0.00\% \\
VGG19 && 80.00\% & 0.00\% && 78.00\% & 0.00\% \\
EfficientNet && 90.00\% & 0.00\% && \underline{98.00\%} & \underline{100.00\%} \\
ConvNext && \underline{92.00\%} & 0.00\% && \underline{98.00\%} & 0.00\% \\
\midrule
ResNet50 & \multirow{7}{5em}{tiger cat} & \textbf{18.33}\% & 0.00\% & \multirow{7}{5em}{persian cat} & 92.00\% & 25.00\%\\
ResNet101 && \textbf{23.33}\% & 0.00\% && 88.00\% & 16.67\%\\
ResNet152 && \textbf{26.67}\% & 0.00\% && 88.00\% & 33.33\%\\
VGG16 && \textbf{20.00}\% & 0.00\% && 86.00\% & 14.29\%\\
VGG19 && \textbf{28.33}\% & 0.00\% && 80.00\% & 10.00\% \\
MobileNetV2 && \textbf{1.67}\% & 0.00\% && \textbf{8.00}\% & 2.17\%\\
EfficientNet && \underline{\textbf{36.67}\%} & 0.00\% && \underline{98.00\%} & \underline{100.00\%}\\
ConvNext && \textbf{26.67}\% & 0.00\% && \underline{98.00\%} & \underline{100.00\%} \\
\midrule
ResNet50 &\multirow{6}{5em}{leopard cat}& \textbf{12.00}\% & \underline{18.18\%} &\multirow{6}{5em}{angora cat} & \textbf{12.00}\% & 50.00\%\\
ResNet101 && \textbf{12.00}\% & 15.91\% && \underline{\textbf{20.00}\%} & 50.00\%\\
ResNet152 && \textbf{4.00}\% & 14.58\% && \underline{\textbf{20.00}\%} & 62.50\%\\
VGG16 && \textbf{10.00}\% & 6.67\% && \textbf{10.00}\% & 46.67\%\\
VGG19 && \textbf{10.00}\% & 6.67\% && \textbf{8.00}\% & 54.35\% \\
EfficientNet && \textbf{2.00}\% & 16.33\% && \textbf{4.00}\% & \underline{95.83\%}\\
ConvNext && \textbf{16.00}\% & 16.67\% && \textbf{10.00}\% & 88.89\%  \\
\midrule
ResNet50 &\multirow{6}{5em}{egyptian cat}& \textbf{24.00}\% & 2.63\% &\multirow{6}{5em}{cat}& 82.05\% & 0.00\%\\
ResNet101 && \textbf{30.00}\% & 2.86\% && 82.05\% & 0.00\% \\
ResNet152 && \textbf{34.00}\% & \underline{6.06}\% && 79.49\% & 0.00\% \\
VGG16 && \textbf{28.00}\% & 0.00\% && 87.18\% & 0.00\% \\
VGG19 && \textbf{26.00}\% & 2.70\% && 76.92\% & 0.00\% \\
MobileNetV2 && \textbf{0.00}\% & 0.00\% && \textbf{2.56}\% & 0.00\%\\
EfficientNet && \underline{\textbf{70.00}\%} & 0.00\% && 92.31\% & 0.00\% \\
ConvNext && \textbf{52.00}\% & 0.00\% && \underline{94.87\%} & 0.00\%\\
\bottomrule
\end{tabular}
\end{table}

\subsection{Convolutional classifiers}
We leveraged the following CNN classifiers: VGG16/19, \cite{vgg}, ResNet50/101/152 \cite{resnet}, InceptionV3 \cite{inception}, InceptionResnetV2 \cite{inceptionresnet}, Xception \cite{Xception}, MobileNetV2 \cite{mobilenet}, NasNet-Large \cite{nasnet}, DenseNet121/169/201 \cite{densenet}, EfficientNet-B7 \cite{efficientnet}, ConvNeXt \cite{convnext}.
We present results for CNN classifiers in Table \ref{tab:cnn-results}. Bold instances denote lower accuracy than the best ImageNet accuracy of each model, as reported by the authors of each model respectively\footnote{\href{https://paperswithcode.com/sota/image-classification-on-imagenet}{https://paperswithcode.com/sota/image-classification-on-imagenet}}. Underlined cells indicate best accuracy/sibling percentage scores for each category.
The absence of models or keywords from Table \ref{tab:cnn-results} means that they correspond to zero accuracy scores. For example, we observe the complete absence of models such as InceptionV3, InceptionResNetV2, Xception, NASNetLarge, DenseNet121/169/201 meaning that they are completely unable to properly classify the crawled images, even those belonging to categories that show satisfactory accuracy when other classifiers are deployed. MobileNetV2 also shows deteriorated performance for all categories.
We will investigate later if hierarchical knowledge can help extract any meaningful information regarding this surprisingly low performance.

Other results that can be extracted from Table \ref{tab:cnn-results} is that some categories can be easily classified ('siamese cat', 'lynx cat', 'cougar cat', 'persian cat', 'cat') contrary to others ('tabby cat', 'tiger cat', 'egyptian cat', 'leopard cat', 'angora cat'). Since we have no specific knowledge of animal species, we will once again leverage WordNet to obtain explanations regarding this behavior. Sibling percentages offer a first glance at the degree of confusion between similar classes in the fine-grained setting. For example, even though 'siamese cat' and 'cougar cat' classes demonstrate high accuracy scores, we observe a completely different behavior regarding the sibling percentages: most CNN classifiers return some sibling false positives for 'siamese cat' ground truth label, while the opposite happens for the 'cougar cat' ground truth label, which mostly receives zero sibling misclassifications. This behavior indicates that for 'siamese cat' if a sample is misclassified, it is likely that it belongs to a conceptually similar class, while for 'cougar cat' misclassifications, false positives belong to more semantically distant categories. 

\begin{table}[htp!]
\caption{Classification results using Transformers. Bold entries denote lower accuracy compared to best model accuracy, underlined metrics indicate best metric performance per class. }
\label{tab:transformer-results}
\begin{tabular}{c|ccc|cccc}
\toprule
Model & Label & Accuracy$\uparrow$ & Siblings$\uparrow$ & Label  & Accuracy$\uparrow$ & Siblings$\uparrow$ \\
\midrule
ViT & \multirow{5}{5em}{tabby cat} & \textbf{44.00}\% & \underline{42.86\%} & \multirow{5}{5em}{siamese cat} & 92.00\% & 50.00\% \\
BeiT & & \textbf{42.00}\% & 3.45\% & & 94.00\% & 0.00\%\\
DeiT & & \underline{\textbf{60.00}\%} & 30.00\% & & 94.00\% & 33.33\% \\
Swin & & \textbf{48.00}\% & 30.77\% & & 94.00\% & \underline{100.00\%}\\
xRegNet & & \textbf{52.00}\% & 25.00\%  & & 92.00\% & 50.00\% \\
CLIP && \textbf{30.00}\% & 28.57\% && \underline{96.00\%} & 50.00\% \\
\midrule
ViT & \multirow{5}{5em}{lynx cat} & 90.00\% & 0.00\% & \multirow{5}{5em}{cougar cat} & \underline{96.00\%} & 0.00\%\\
BeiT & & \textbf{26.00}\% & 0.00\% & & 92.00\% & 0.00\%\\
DeiT & & \underline{92.00\%} & 0.00\%  & & \underline{96.00\%} & 0.00\% \\
Swin & & 86.00\% & 0.00\% & & \underline{96.00\%} & 0.00\% \\
xRegNet &  & 90.00\% & 0.00\% & & \underline{96.00\%} & \underline{50.00\%} \\
CLIP && 86.00\% & 0.00\% && 92.00\% & 0.00\% \\
\midrule
ViT & \multirow{5}{5em}{tiger cat} & \textbf{18.33}\% & 0.00\% & \multirow{5}{5em}{persian cat} & 92.00\% & 75.00\%\\
BeiT &  & \textbf{18.33}\% & \underline{22.45\%} & & 80.00\% & 10.00\% \\
DeiT &  & \textbf{15.00}\% & 0.00\% &  & 96.00\% & 50.00\%\\
Swin &  & \textbf{21.67}\% & 0.00\% &  & 96.00\% & 50.00\%\\
xRegNet & & \underline{\textbf{35.00}\%} & 0.00\% &  & \underline{98.00\%} & \underline{100.00\%}\\
CLIP && 46.67\% & 0.00\% && 96.00\% & 50.00\%\\
\midrule
ViT & \multirow{5}{5em}{leopard cat} & \textbf{12.00}\% & 2.27\% & \multirow{5}{5em}{angora cat}  & \textbf{6.00}\% & 89.36\%\\
BeiT &  & \textbf{6.00}\% & 78.72\% & & \underline{\textbf{62.00}\%} & 52.63\% \\
DeiT &  & \textbf{10.00}\% & 11.11\% & & \textbf{0.00}\% & 94.00\%\\
Swin &  & \underline{\textbf{14.00}\%} & 9.30\% & & \textbf{8.00}\% & \underline{95.65\%}\\
xRegNet &  & \textbf{6.00}\% & 21.28\% & & \textbf{0.00}\% & 76.00\%\\
CLIP &  & \textbf{10.00}\% & \underline{55.56\%} & & \textbf{8.00}\% & 91.30\% \\
\midrule
ViT & \multirow{5}{5em}{egyptian cat} & \textbf{38.00}\% & 3.23\% & \multirow{5}{5em}{cat} & 89.74\% & 0.00\% \\
BeiT & & \textbf{20.00}\% & \underline{22.50\%} &  & \textbf{53.85}\% & 0.00\%\\
DeiT & & \textbf{36.00}\% & 3.12\%  &  & \underline{94.87\%} & 0.00\%\\
Swin & & \underline{\textbf{52.00}\%} & 0.00\% &  & 89.74\% & 0.00\%\\
xRegNet && \textbf{48.00}\% & 0.00\% &  & 92.31\% & 0.00\% \\
CLIP && 70.00\% & 6.67\% && 69.23\% & 0.00\% \\
\bottomrule
\end{tabular}
\end{table}

Regarding model capabilities, we observe that for both 'siamese' and 'cougar cat' classes, all ResNet50 false positives belong to non-sibling classes, contrary to EfficientNet false positives, which all belong to sibling classes. By also looking to other categories, we observe that in general, EfficientNet achieves a higher sibling percentage compared to ResNet50, meaning that EfficientNet misclassifications are more justified compared to ResNet50 misclassifications.

\subsection{Transformer-based classifiers}
The following transformer-based image classifiers were used: ViT  \cite{vit}, Regnet-x \cite{regnet}, DeiT \cite{deit}, BeiT \cite{beit}, CLIP \cite{clip}, Swin Transformer V2 \cite{swin}.
Results for Transformer-based classifiers are provided in Table \ref{tab:transformer-results}. We spot a similar pattern regarding the categories upon which models struggle to make predictions: instances belonging to 'tabby cat', 'tiger cat', 'egyptian cat' categories are classified with low accuracy compared to 'siamese cat', 'lynx cat', 'cougar cat', 'persian cat', 'cat', 'angora cat' and 'leopard cat'. We suspect that there is a common reason behind this behavior, probably attributed to unavoidable intra-class similarities present in the fine-grained classification setting. 

As for model performance, we examine sibling percentage apart from exclusively evaluating accuracy. The behavior of transformer-based models regarding sibling misclassification is harder to be interpreted compared to CNN models, because models that return high sibling percentages for some categories may present low sibling percentages on other categories and vice versa. For example, BeiT scores low on sibling percentages for 'tabby cat' (3.45\%), 'siamese cat' (0\%) and 'persian cat' (10\%) compared to other models for the same classes; on the other hand, it returns \textit{best} sibling scores for 'leopard cat' (78.72\%), 'tiger cat' (22.45\%) and 'egyptian cat' (22.50\%).
More results about the explainability of results are provided in Section \ref{sec:explain}.

\subsection{Explaining misconceptions}
\label{sec:explain}

\begin{table}[htp!]
\caption{Common misclassifications for selected GT cat classes and misclassification frequency (CNNs).}
\label{tab:miscl-cnn}
\begin{tabular}{>{\centering\arraybackslash}p{2.1em}|>
{\centering\arraybackslash}p{3em}|>{\centering\arraybackslash}p{7.5em}>{\centering\arraybackslash}p{2.3em}|>{\centering\arraybackslash}p{7.5em}>{\centering\arraybackslash}p{2.3em}|>{\centering\arraybackslash}p{5.9em}>{\centering\arraybackslash}p{2em}}
\toprule
&  & \multicolumn{2}{c}{Top-1} &  \multicolumn{2}{c}{Top-2} &  \multicolumn{2}{c}{Top-3} \\
\cline{3-8}
Model & GT & FP & MF & FP & MF & FP & MF \\
\toprule
\multirow{10}{3em}{Res\\Net50} & tabby & \textcolor{blue}{tiger cat} & 32.00\% &
\textcolor{blue}{egyptian cat} & 24.00\%& \textcolor{red}{web site} & 8.00\\
& angora & \textcolor{blue}{persian cat} & 34.00\% & \textcolor{red}{arctic fox} & 11.36\% & lynx & 9.09\%\\
&lynx & \textcolor{red}{coyote} & 22.22\% & tabby cat & 11.11\% & egyptian cat & 11.11\%\\
& siamese & \textcolor{red}{great dane} & 20.00\% & \textcolor{red}{hare} & 20.00\% & \textcolor{red}{american egret} & 20.00\%\\
& tiger& \textcolor{blue}{tabby cat} & 40.82\% & \textcolor{blue}{egyptian cat} & 20.41\% & tiger & 14.29\%\\
& persian & \textcolor{red}{old English sheepdog} & 25.00\% & \textcolor{blue}{siamese cat} & 25.00\% & \textcolor{red}{hatchet} & 25.00\%\\
& cougar & \textcolor{blue}{lynx} & 25.00\% & \textcolor{red}{malinois} & 25.00\% & \textcolor{red}{wallaby} & 25.00\%\\
&leopard & egyptian cat & 30.00\% & \textcolor{blue}{tiger cat} & 16.00\% &
 jaguar & 12.00\%\\
& egyptian & \textcolor{red}{mexican hairless} & 10.53\% & \textcolor{red}{mask} & 5.26\% & \textcolor{red}{comic book} & 5.26\%\\
& cat & \textcolor{red}{fur coat} & 14.29\% & \textcolor{red}{carton} & 14.29\% & \textcolor{red}{book jacket} & 14.29\%\\
\midrule
\multirow{10}{5em}{Res\\Net\\101} & tabby & \textcolor{blue}{egyptian cat} & 41.67\% & \textcolor{blue}{tiger cat} & 29.17\% & \textcolor{red}{web site} & 8.33\%\\
& angora & \textcolor{blue}{persian cat} & 32.50\% & \textcolor{blue}{egyptian cat} & 12.50 & lynx & 10.00\% \\
& lynx & tabby cat & 12.50\% & egyptian cat & 12.50\% & cheetah & 12.50\% \\
& siamese & \textcolor{red}{Boston bull} & 16.67\% & \textcolor{blue}{egyptian cat} & 16.67\% & \textcolor{red}{hare} & 16.67\%\\
& tiger & \textcolor{blue}{tabby cat} & 34.78\% & \textcolor{blue}{tiger cat} & 17.39\% & \textcolor{blue}{egyptian cat} & 15.22\% \\
& persian & \textcolor{red}{keeshond} & 16.67\% & \textcolor{red}{guinea pig} & 16.67\% & \textcolor{red}{collie} & 16.67\% \\
& cougar & lynx & 45.45\% & \textcolor{red}{meerkat} & 9.09\%& \textcolor{red}{dhole} & 9.09\% \\
& leopard & egyptian cat & 36.00\% & \textcolor{blue}{tiger cat} & 14.00\% & leopard & 12.00\% \\
& egyptian & \textcolor{red}{mexican hairless} & 14.29\% & \textcolor{red}{mask} & 8.57\% & \textcolor{red}{sea lion}  & 5.71\% \\
& cat & \textcolor{red}{macaque} & 14.29\% & \textcolor{red}{barbershop} & 14.29\% & \textcolor{red}{Pembroke} & 14.29\% \\
\midrule
\multirow{12}{1em}{Res\\Net\\152} & tabby & \textcolor{blue}{tiger cat} & 40.00\% & \textcolor{blue}{egyptian cat} & 12.00\% & lynx & 12.00\% \\
& angora & \textcolor{blue}{persian cat} & 35.00\% & \textcolor{blue}{siamese cat} & 10.00\% & \textcolor{red}{shower curtain } & 10.00\% \\
& lynx & tabby cat & 42.86\% & \textcolor{red}{coyote} & 14.29\% & \textcolor{red}{norwich terrier} & 14.29\\
& siamese & \textcolor{red}{whippet} & 20.00\% & \textcolor{blue}{egyptian cat} & 20.00\% & \textcolor{blue}{angora cat} & 20.00\% \\
& tiger & \textcolor{blue}{tabby cat} & 34.09\% & \textcolor{blue}{egyptian cat} & 18.18\% & tiger & 15.91\% \\
& persian & \textcolor{blue}{siamese cat} & 33.33\%  & \textcolor{red}{collie} & 16.67\% & \textcolor{red}{fur coat} & 16.67\%\\
& cougar & \textcolor{red}{menu} & 16.67\% & \textcolor{red}{wild boar} & 16.67\% & \textcolor{red}{wallaby} & 16.67\%\\
& leopard & egyptian cat & 28.00\% & \textcolor{blue}{lynx} & 22.00\% & jaguar & 16.00\% \\
& egyptian & \textcolor{red}{mexican hairless} & 18.18\% & \textcolor{red}{web site} & 9.09\% & \textcolor{blue}{tabby cat} & 6.06\% \\
& cat & \textcolor{red}{macaque} & 12.50\% & \textcolor{red}{Pembroke} & 12.50\% & \textcolor{red}{chihuahua} & 12.50\%\\
\midrule
\multirow{10}{1em}{VGG\\16} & tabby & \textcolor{blue}{egyptian cat} & 38.71\% & \textcolor{blue}{tiger cat} & 22.58\% & \textcolor{red}{wood rabbit} & 3.23\%\\
& angora & \textcolor{blue}{persian cat} & 26.67\% & \textcolor{blue}{egyptian cat} & 15.56\% & lynx & 8.89\% \\
& lynx & \textcolor{red}{coyote} & 33.33 & egyptian cat & 22.22\% & madagascar & 11.11\% \\
& siamese & \textcolor{red}{mexican hairless} & 22.22\%  & \textcolor{red}{whippet} & 11.11\% & \textcolor{red}{fur coat} & 11.11\%\\
& tiger & \textcolor{blue}{tabby cat} & 33.33\% & \textcolor{blue}{egyptian cat} & 20.83\% & tiger & 16.67\% \\
& persian & \textcolor{red}{arctic fox} & 14.29\% & \textcolor{blue}{angora cat} & 14.29\% & lynx & 14.29\%\\
& cougar & \textcolor{blue}{lynx} & 42.86\% & \textcolor{red}{coyote} & 28.57\%& \textcolor{red}{menu} & 14.29\%\\
& leopard & egyptian cat & 42.00\% & \textcolor{blue}{lynx} & 18.00\% & jaguar & 10.00\% \\
& egyptian  & \textcolor{red}{mexican hairless} & 8.33\% &  lynx & 5.56\% & \textcolor{red}{sombrero } & 5.56\%\\
& cat & \textcolor{red}{norwich terrier} & 20.00\% & \textcolor{red}{schipperke} & 20.00\% & \textcolor{red}{kit fox} & 20.00\%\\
\bottomrule
\end{tabular}
\end{table}

In Tables \ref{tab:miscl-cnn},\ref{tab:miscl} \& \ref{tab:miscl2} we report the top-3 misclassifications per ground truth (GT) category and per model, as well as the misclassification frequency (MF) for each false positive (FP) label. GT column refers to cat species exclusively, even if the word 'cat' is omitted (for example, 'tiger' GT entry refers to 'tiger cat').
We highlight with \textcolor{red}{red} irrelevant FP classes, which are semantically distant compared to the GT label, while misconceptions involving sibling classes are highlighted with \textcolor{blue}{blue}. Moreover, \textcolor{magenta}{magenta} indicates that an FP is actually an immediate (1 hop) hypernym of the GT.
Due to space constraints, we present here all transformer-based models, but only a subset of the CNN models tested in total; more results can be found in the Appendix.

\begin{table}[htp!]
\caption{Common misclassifications for selected GT cat classes and misclassification frequency (Transformers).}
\label{tab:miscl}
\begin{tabular}{>{\centering\arraybackslash}p{2.2em}|>
{\centering\arraybackslash}p{3em}|>{\centering\arraybackslash}p{7.5em}>{\centering\arraybackslash}p{2.3em}|>{\centering\arraybackslash}p{7.5em}>{\centering\arraybackslash}p{2.3em}|>{\centering\arraybackslash}p{5.8em}>{\centering\arraybackslash}p{2em}}
\toprule
&  & \multicolumn{2}{c}{Top-1} &  \multicolumn{2}{c}{Top-2} &  \multicolumn{2}{c}{Top-3} \\
\cline{3-8}
Model & GT & FP & MF & FP & MF & FP & MF \\
\toprule
\multirow{10}{1em}{CLIP} & tabby & madagascar & 40.00\% &
\textcolor{blue}{egyptian cat} & 22.86\%& \textcolor{blue}{tiger cat} & 11.43\\
& angora & \textcolor{blue}{persian cat} & 78.26\% & madagascar & 6.52\% & \textcolor{blue}{siamese cat} & 6.52\%\\
&lynx & madagascar & 14.29\% & \textcolor{blue}{leopard cat} & 14.29\% & \textcolor{red}{grey fox} & 14.29\%\\
& siamese & \textcolor{red}{polecat} & 50.00\% & \textcolor{blue}{persian cat} & 50.00\% & - & -\\
& tiger & \textcolor{blue}{egyptian cat} & 30.77\% & madagascar & 19.23\% & \textcolor{blue}{leopard cat} & 15.38\%\\
& persian & madagascar & 50.00\% & \textcolor{blue}{siamese cat} & 50.00\%&-&-\\
& cougar & \textcolor{blue}{lynx} & 75.00\% & madagascar & 25.00\%&-&-\\
&leopard & \textcolor{blue}{tiger cat} & 55.56\% & madagascar & 17.78\%
 & egyptian cat & 11.11\%\\
& egyptian & \textcolor{red}{mexican hairless} & 26.67\% & madagascar & 26.67\% & \textcolor{red}{armadillo} & 6.67\%\\
& cat & madagascar & 66.67\% & orange & 16.67\% & bib & 8.33\%\\
\midrule
\multirow{10}{1em}{BeiT} & tabby & \textcolor{blue}{tiger cat} & 17.24\% & cat & 13.79\% & \textcolor{magenta}{domestic} & 13.79\%\\
& angora & \textcolor{blue}{persian cat} & 42.11\% & \textcolor{magenta}{domestic} & 21.05\% & quadruped & 5.26\% \\
& lynx & \textcolor{blue}{common lynx} & 59.46\% & \textcolor{blue}{Canada lynx} & 16.22\% & bobcat & 5.41\% \\
& siamese & \textcolor{blue}{kitten} & 66.67\% & feline & 33.33\%&-&-\\
& tiger & \textcolor{blue}{tabby cat} & 23.40\% & \textcolor{blue}{margay} & 14.89\% & \textcolor{magenta}{domestic} & 6.38\% \\
& persian & \textcolor{magenta}{domestic} & 20.00\% & \textcolor{blue}{angora cat} & 10.00\% & \textcolor{red}{breadwinner} & 10.00\% \\
& cougar & feline & 25.00\% & big cat & 25.00\% & cub & 25.00\% \\
& leopard & \textcolor{blue}{margay} & 42.55\% & \textcolor{blue}{ocelot} & 21.28\% & spotted lynx & 8.51\% \\
& egyptian & \textcolor{blue}{Abyssinian} & 15.00\% & \textcolor{red}{mexican hairless} & 10.00\% & \textcolor{blue}{mouser} & 5.00\% \\
& cat & \textcolor{magenta}{feline} & 33.33\% & kitten & 22.22\% & caterer & 11.11\% \\
\midrule
\multirow{12}{1em}{DeiT} & tabby & \textcolor{blue}{tiger cat} & 35.00\% & \textcolor{blue}{egyptian cat} & 30.00\% & \textcolor{red}{web site} & 15.00\% \\
& angora & \textcolor{blue}{persian cat} & 62.00\% & \textcolor{blue}{egyptian cat} & 28.00\% & \textcolor{blue}{tabby cat} & 2.00\% \\
& lynx & tabby cat & 75.00\% & \textcolor{red}{coyote} & 25.00\% &-&-\\
& siamese & \textcolor{blue}{egyptian cat} & 33.33\% & \textcolor{red}{mexican hairless} & 33.33\% & lynx & 33.33\% \\
& tiger & \textcolor{blue}{tabby cat} & 37.50\% & \textcolor{blue}{egyptian cat} & 27.50\% & \textcolor{blue}{leopard cat} & 12.50\% \\
& persian & \textcolor{red}{soft-coated wheaten terrier} & 50.00\%  & \textcolor{blue}{siamese cat} & 50.00\% &-&-\\
& cougar & \textcolor{red}{web site} & 50.00\% & \textcolor{red}{dingo} & 50.00\%&-&-\\
& leopard & egyptian cat & 48.89\% & \textcolor{blue}{lynx} & 22.22\% & \textcolor{blue}{tiger cat} & 11.11\% \\
& egyptian & \textcolor{red}{mexican hairless} & 15.62\% & \textcolor{red}{comic book} & 9.38\% & \textcolor{red}{kelpie} & 3.12\% \\
& cat & \textcolor{red}{fur coat} & 50.00\% & \textcolor{red}{chihuahua} & 50.00\%&-&-\\
\midrule
\multirow{10}{1em}{xReg\\Net} & tabby & \textcolor{blue}{tiger cat} & 62.50\% & \textcolor{blue}{egyptian cat} & 20.83\% & \textcolor{red}{menu} & 4.17\%\\
& angora & \textcolor{blue}{persian cat} & 48.00\% & \textcolor{blue}{egyptian cat} & 18.00\% & lynx & 8.00\% \\
& lynx & tabby cat & 40.00\% & \textcolor{blue}{tiger cat} & 20.00\% & egyptian cat & 20.00\% \\
& siamese & \textcolor{blue}{egyptian cat} & 50.00\%  & \textcolor{red}{polecat} & 25.00\%
 & lynx & 25.00\% \\
 & tiger & \textcolor{blue}{tabby cat} & 40.00\% & \textcolor{blue}{egyptian cat} & 25.71\% & \textcolor{blue}{lynx} & 11.43\% \\
& persian & \textcolor{blue}{siamese cat} & 100.0\% &-&-&-&-\\
& cougar & \textcolor{blue}{tiger cat} & 50.00\% & \textcolor{blue}{lynx} & 50.00\%&-&-\\
& leopard & egyptian cat & 40.43\% & \textcolor{blue}{tiger cat} & 21.28\% & \textcolor{blue}{lynx} & 17.02\% \\
& egyptian & \textcolor{red}{mexican hairless} & 15.38\% & \textcolor{red}{mask} & 7.69\% & \textcolor{red}{comic book} & 7.69\%\\
& cat & \textcolor{red}{comic book} & 33.33\% & \textcolor{red}{tub} & 33.33\% & \textcolor{red}{drake} & 33.33\% \\
\bottomrule
\end{tabular}
\end{table}

\begin{table}[htp!]
\caption{(Continuation of Tab \ref{tab:miscl}). Common misclassifications and misclassification frequency.}
\hspace{-6px}
\label{tab:miscl2}
\begin{tabular}{>{\arraybackslash}p{2.1em}|>
{\centering\arraybackslash}p{3em}|>
{\centering\arraybackslash}p{7.5em}>{\centering\arraybackslash}p{2.2em}|>{\centering\arraybackslash}p{6.1em}>{\centering\arraybackslash}p{2.2em}|>{\centering\arraybackslash}p{7.5em}>{\centering\arraybackslash}p{2em}}
\toprule
&  & \multicolumn{2}{c}{Top-1} &  \multicolumn{2}{c}{Top-2} &  \multicolumn{2}{c}{Top-3} \\
\cline{3-8}
Model & GT & FP & MF & FP & MF & FP & MF \\
\toprule
\multirow{10}{0.5em}{Swin} & tabby & \textcolor{blue}{tiger cat} & 57.69\% & \textcolor{blue}{egyptian cat} & 30.77\% & \textcolor{red}{web site} & 7.69\% \\
& angora & \textcolor{blue}{persian cat} & 58.70\% & \textcolor{blue}{egyptian cat} & 26.09\% & \textcolor{blue}{tabby cat} & 10.87\% \\
& lynx & tabby cat & 57.14\% & \textcolor{red}{fur coat} & 14.29\% & \textcolor{red}{timber wolf} & 14.29\% \\
& siamese & \textcolor{blue}{egyptian cat} & 100.0\%&-&-&-&-\\
& tiger & \textcolor{blue}{tabby cat} & 35.00\% & \textcolor{blue}{egyptian cat} & 32.50\% & \textcolor{blue}{leopard cat} & 12.50\% \\
& persian & \textcolor{blue}{siamese cat} & 50.00\% & \textcolor{red}{hand blower} & 50.00\%&-&-\\
& cougar & \textcolor{red}{web site} & 50.00\% & \textcolor{red}{Irish wolfhound} & 50.00\%&-&-\\
& leopard & egyptian cat & 44.19\% & \textcolor{blue}{lynx} & 37.21\% & \textcolor{blue}{tiger cat} & 9.30\% \\
& egyptian & \textcolor{red}{mexican hairless} & 20.83\% & \textcolor{red}{comic book} & 16.67\% & \textcolor{red}{table lamp} & 8.33\% \\
& cat & \textcolor{red}{fur coat} & 25.00\% & \textcolor{red}{jersey} & 25.00\% & \textcolor{red}{chihuahua} & 25.00\% \\
\midrule
\multirow{10}{0.5em}{ViT} & tabby & \textcolor{blue}{egyptian cat} & 42.86\% & \textcolor{blue}{tiger cat} & 32.14\% & \textcolor{red}{web site} & 10.71\%\\
& angora & \textcolor{blue}{egyptian cat} & 48.94\% & \textcolor{blue}{persian cat} & 38.30\% & \textcolor{blue}{tabby cat} & 2.13\% \\
& lynx & tabby cat & 40.00\% & egyptian cat & 40.00\% & \textcolor{red}{timber wolf} & 20.00\% \\
& siamese & \textcolor{blue}{egyptian cat} & 50.00\% & \textcolor{red}{chihuahua} & 25.00\% & \textcolor{red}{mexican hairless} & 25.00\% \\
& tiger & \textcolor{blue}{egyptian cat} & 48.78\% & \textcolor{blue}{tabby cat} & 26.83\% & \textcolor{blue}{leopard cat} & 14.63\% \\
& persian & \textcolor{red}{plastic bag} & 25.00\% & \textcolor{blue}{egyptian cat} & 25.00\% & \textcolor{blue}{siamese cat} & 25.00\% \\
& cougar & egyptian cat & 50.00\% & \textcolor{red}{malinois} & 50.00\%&-&-\\
& leopard & egyptian cat & 86.36\% & snow leopard & 2.27\% & \textcolor{red}{web site} & 2.27\% \\
& egyptian & \textcolor{red}{mexican hairless} & 16.13\% & \textcolor{red}{pedestal} & 12.90\% & \textcolor{red}{vase} & 6.45\% \\
& cat & \textcolor{red}{washer} & 25.00\% & \textcolor{red}{fur coat} & 25.00\% & \textcolor{red}{mexican hairless} & 25.00\% \\
\bottomrule
\end{tabular}
\end{table}

Interestingly, we can spot some surprising frequent misconceptions, such as confusing cat species with the 'mexican hairless' dog breed. 
For CNN classifiers, we spot this peculiarity for all models under investigation: 10.53\% of ResNet50 FP for 'egyptian cat' GT label belong to the 'mexican hairless' class; the same applies to 14.29\% of ResNet101 FP, 18.18\% of ResNet152 FP and 8.33\% of VGG16 FP. More animals such as 'wallaby', 'jaguar', 'sea lion', 'cheetah', 'arctic fox', 'coyote' etc appear as frequent FPs.

For transformer models, the 'egyptian cat' $\rightarrow$  'mexican hairless' abnormality is observed for all classifiers when 'egyptian cat' GT label is provided, resulting in the following 'mexican hairless' FP percentages: 26.67\% for CLIP, 10\% for BeiT, 15.62\% for DeiT, 15.38\% for xRegNet, 20.83\% for Swin, and 16.33\% for ViT. Obviously, regardless of whether the CNN or transformer classifier is being used, images of 'egyptian cats' are often erroneously perceived as 'mexican hairless dogs'. A qualitative analysis between 'egyptian cat' images and 'mexican hairless dog' images indicates that these animals are obviously distinct, even though they present similar ear shapes and rather hairless, thin bodies. Therefore, we can assume that the transformer-based classifiers are biased towards texture, verifying relevant observations reported for CNNs \cite{robustness5}. Also, ear shape acts as a confounding factor, overshadowing other actually distinct animal characteristics. There are more misclassifications involving animals, such as 'armadillo', 'chihuahua', 'soft-coated wheaten terrier', 'kelpie', and others.

Even more surprising are misclassifications not including animal species. For example, CNN classifiers predict 'web site' instead of 'tabby cat', 'hatched' instead of 'persian cat', 'barbershop' instead of 'cat', 'menu' instead of 'cougar' etc. All ResNet50/101/152 and VGG16 make at least one such misclassification, something that highly questions which features of cat species contribute to such predictions.

Misclassifications involving non-animal classes using transformers (Tables \ref{tab:miscl}, \ref{tab:miscl2}) provide the following interesting abnormalities: 'cat' is classified as 'fur coat' for 50\% of the FP instances using DeiT. This non-negligible misclassification rate once again verifies the aforementioned texture bias. In a similar sense, xRegNet classifies 'egyptian cat' images as 'mask' and as 'comic book'  7.69\% of the FPs respectively. Such categories had also appeared in CNN misclassifications.
We cannot provide a human-interpretable explanation about the 'mask' misclassification, since the term 'mask' may refer to many different objects. We hypothesize that 'mask' ImageNet instances may contain carnival masks looking similar to cats, therefore the lack of context confused xRegNet. 'Comic book' appears 9.38\% of the times an 'egyptian cat' image is misclassified by DeiT, 33.33\% of the times a 'cat' photo is misclassified by xRegNet, and 16.67\% of the times an 'egyptian cat' is misclassified by Swin. This can be attributed to the fact that crawled images may contain cartoon-like instances, which cannot be clearly regarded as cats.
Other interesting misclassifications involving irrelevant categories are 'cat'$\rightarrow$'washer' (25\% of FPs using ViT), 'leopard cat'$\rightarrow$'web site' (2.27\% of FPs using ViT, 15\% of FPs using DeiT), 'persian cat'$\rightarrow$'plastic bag' (25\% of FPs using ViT), 'cat'$\rightarrow$'jersey' (25\% of FPs using Swin), 'egyptian cat'$\rightarrow$'table lamp' (8.33\% of FPs using Swin), 'cat'$\rightarrow$'tub (33.33\% of FPs using xRegNet), and others. 

An interesting observation revolves around the 'egyptian cat' label. For CNN models, almost all top-3 FP of 'egyptian cat' GT label correspond to irrelevant ImageNet categories. On the contrary, 'tabby cat', 'angora cat', and 'tiger cat' present more sensible FPs, which usually involve sibling categories (highlighted with blue).
As for transformer models, we observe that 'egyptian cat' label is always being confused with at least one irrelevant ImageNet category, while 'angora cat' is only confused with other cat species, and not with conceptually distant classes. Thus, 'egyptian cat' crawled images seem to contain some misleading visual features that frequently derail the classification process. Indeed, when viewing 'egyptian cat' crawled images, some of them are drawings or photos of cat souvenirs; however, misconceptions such as 'table lamp' or 'armadillo' cannot be visually explained by human inspectors, unraveling more questions on the topic.
A comparison between CNN classifiers (Table \ref{tab:miscl-cnn}) and transformer-based classifiers (Table \ref{tab:miscl}, \ref{tab:miscl2}) denotes that transformers are more capable of retrieving similar categories to the GT; this becomes obvious by observing the higher number or irrelevant misclassifications highlighted with red for CNNs, compared to transformer results. 

By combining Tables \ref{tab:miscl-cnn}, \ref{tab:miscl} \& \ref{tab:miscl2} with Tables \ref{tab:cnn-results}\& \ref{tab:transformer-results}, we obtain some very interesting findings: how are low classification metric scores connected to the relevance between misclassified categories? 
We start with categories presenting low accuracy scores ('tabby cat', 'tiger cat', 'egyptian cat'), and we compare them with categories offering frequent extraneous misclassifications ('egyptian cat' and 'cat', followed by 'tabby cat' and 'lynx'). Classifying 'egyptian cat' images both yields low classification scores and returns irrelevant false positives. On the other hand, even though 'cat' images present high accuracy scores, misclassifications are highly unrelated when they happen. 'Tiger cat' scores low in accuracy, however, misclassifications are rather justified, since other cat species are returned. Surprisingly, 'tiger cat' also scores low in siblings percentage, indicating that false positives are not immediately related to the GT 'tiger cat' class. In this case, we assume that false positives ('egyptian cat', 'tabby cat', 'leopard cat' etc) belong to more distant relatives of the 'tiger cat' concept class, even though bearing some similar features.

\begin{table}[b]
\caption{Conceptual metrics based on WordNet distances using CNN classifiers.}
\label{tab:kdm}
\begin{tabular}{>{\arraybackslash}p{5em}|>{\arraybackslash}p{3em}ccc|>{\arraybackslash}p{3em}ccc}
\toprule
Model & Label & Path sim$\uparrow$ & LCH$\uparrow$ & WUPS$\uparrow$  & Label & Path sim$\uparrow$ & LCH$\uparrow$ & WUPS$\uparrow$ \\
\toprule
ResNet50 &\multirow{6}{4em}{tabby \\ cat}& 0.18 & 1.79 & 0.69 & \multirow{6}{4em}{siamese cat}& 0.10 & 1.25 & 0.57 \\
ResNet101 && 0.22 & 1.99 & 0.75 && 0.15 & 1.60 & 0.72\\
ResNet152 &&0.16 & 1.59 & 0.62 && 0.17 & 1.71 & 0.67\\
VGG16 && 0.21 & 1.85 & 0.70 && 0.16 & 1.65 & 0.65\\
VGG19 && 0.18 & 1.68 & 0.63 && 0.17 & 1.73 & 0.70\\
MobileNetV2 && 0.09 & 1.13 & 0.39 && 0.09 & 1.15 & 0.40\\
EfficientNet && \textbf{0.24} & \textbf{2.17} & \textbf{0.86} && \textbf{0.33} & \textbf{2.54} & \textbf{0.88}\\
\midrule
ResNet50 &\multirow{6}{4em}{lynx cat}& 0.05 & 0.53 & 0.11 &\multirow{6}{4em}{cougar cat}& 0.08 & 0.97 & 0.49\\
ResNet101 &&0.05 & \textbf{0.62} & \textbf{0.13} && 0.08 & 0.90 & 0.41 \\
ResNet152 && 0.05 & 0.56 & 0.09 && 0.08 & 1.01 & 0.49\\
VGG16 && 0.04 & 0.46 & 0.08 && 0.08 & 0.88 & 0.37 \\
VGG19 && 0.04 & 0.48 & 0.08  && 0.07 & 0.86 & 0.38\\
EfficientNet && 0.05 & 0.54 & 0.09  && \textbf{0.33} & \textbf{2.54} & \textbf{0.94}\\
\midrule
ResNet50 &\multirow{6}{5em}{tiger cat}& \textbf{0.15} & \textbf{1.61} & \textbf{0.70} &\multirow{6}{5em}{persian \\ cat}& 0.16 & 1.59 & 0.60\\
ResNet101 && 0.14 & 1.47 & 0.64 && 0.17 & 1.73 & 0.66\\
ResNet152 && 0.13 & 1.43 & 0.61 && 0.17 & 1.57 & 0.56\\
VGG16 && 0.14 & 1.51 & 0.65 && 0.12 & 1.27 & 0.50\\
VGG19 && 0.13 & 1.43 & 0.61 && 0.13 & 1.45 & 0.57 \\
MobileNetV2 && 0.07 & 0.90 & 0.41 && 0.09 & 1.19 & 0.43\\
EfficientNet && 0.13 & 1.41 & 0.60 && \textbf{0.33} & \textbf{2.54} & \textbf{0.88}\\
\midrule
ResNet50 &\multirow{6}{5em}{leopard \\ cat}& 0.17 & 1.62 & 0.67 &\multirow{6}{5em}{angora \\ cat}& 0.22 & 1.94 & 0.72 \\
ResNet101 && 0.17 & 1.62 & 0.67 && 0.22 & 1.93 & 0.71\\
ResNet152 && 0.16 & 1.55 & 0.64 && 0.24 & 2.05 & 0.73\\
VGG16 && 0.15 & 1.49 & 0.62 && 0.21 & 1.90 & 0.72\\
VGG19 && 0.15 & 1.53 & 0.64 && 0.23 & 2.01 & 0.75\\
EfficientNet && \textbf{0.18} & \textbf{1.68} & \textbf{0.68} && \textbf{0.32} & \textbf{2.48} & \textbf{0.86} \\
\midrule
ResNet50 &\multirow{6}{5em}{egyptian \\ cat}& 0.11 & 1.32 & 0.51 &\multirow{6}{5em}{cat}& 0.11 & 1.29 & 0.56\\
ResNet101 && 0.11 & 1.32 & 0.50 && 0.11 & 1.34 & 0.63\\
ResNet152 && \textbf{0.12} & \textbf{1.40} & \textbf{0.55} && 0.11 & 1.35 & 0.61\\
VGG16 && 0.09 & 1.15 & 0.41 && \textbf{0.14} & \textbf{1.67} & \textbf{0.80}\\
VGG19 && 0.10 & 1.21 & 0.45 && 0.12 & 1.42 & 0.64 \\
MobileNetV2 && 0.08 & 1.06 & 0.36 && 0.07 & 0.88 & 0.34 \\
EfficientNet && 0.10 & 1.24 & 0.49 && 0.12 & 1.52 & 0.74\\
\bottomrule
\end{tabular}
\end{table}

Overall, throughout this analysis we prove that classification accuracy is unable to reveal the whole truth behind the way classifiers behave; to this end, knowledge sources are able to shed some light on the inner workings of this process. By analyzing a constraint family of related ImageNet labels (cat species) we already disentangled the classification accuracy from the classification \textit{relevance}: false positives can be highly relevant to the ground truth (such as 'tiger cat' misclassifications) or not ('cat' misclassifications). We, therefore, argue that fine-grained classification also demands \textit{fine-grained evaluation}, which can provide insightful information when driven by knowledge. The human interpretable insights of Tables \ref{tab:miscl}, \ref{tab:miscl2} are going to be quantified and verified in the next Section.

\subsection{Knowledge-driven metrics}

\begin{table}[b!]
\caption{(Continuation of Tab \ref{tab:kdm}). Conceptual metrics based on WordNet distances using transformers.}
\label{tab:kdm2}
\begin{tabular}{>{\arraybackslash}p{3.3em}|>{\arraybackslash}p{3em}ccc|>{\arraybackslash}p{3em}ccc}
\toprule
Model & Label & Path sim$\uparrow$ & LCH$\uparrow$ & WUPS$\uparrow$ & Label & Path sim$\uparrow$ & LCH$\uparrow$ & WUPS$\uparrow$ \\
\midrule
ViT & \multirow{6}{5em}{tabby \\ cat} & 0.23 & 2.02 & 0.76 & \multirow{6}{5em}{siamese \\ cat} & 0.23 & 2.01 & 0.71 \\
BeiT &  & \textbf{0.24} & 2.02 & 0.75 & & 0.18 & 1.88 & 0.77\\
DeiT &  & 0.20 & 1.83 & 0.70 && 0.19 & 1.72 & 0.58 \\
Swin &  & 0.23 & \textbf{2.07} & \textbf{0.82} && \textbf{0.33} & \textbf{2.54} & \textbf{0.88} \\
xRegNet &  & 0.22 & 2.00 & 0.79 && 0.21 & 1.84 & 0.66\\
CLIP &  & 0.18 & 1.78 & 0.75 && 0.24 & 2.12 & 0.84\\
\midrule
ViT & \multirow{6}{5em}{lynx cat} & 0.05 & 0.55 & 0.09 & \multirow{6}{5em}{cougar \\ cat} & 0.15 & 1.63 & \textbf{0.79}\\
BeiT & & 0.04 & 0.33 & 0.07 && \textbf{0.21} & \textbf{1.94} & \textbf{0.79}\\
DeiT &  & 0.05 & 0.54 & 0.09 && 0.09 & 1.13 & 0.54\\
Swin & & 0.05 & \textbf{0.56} & 0.09 && 0.07 & 0.97 & 0.51\\
xRegNet & & 0.04 & 0.50 & 0.08 && 0.19 & 1.44 & 0.50\\
CLIP & & 0.04 & 0.51 & \textbf{0.10} && 0.06 & 0.62 & 0.24\\
\midrule
ViT & \multirow{6}{5em}{tiger cat} & 0.15 & 1.60 & 0.69 & \multirow{6}{5em}{persian \\ cat} & 0.27 & 2.19 & 0.76\\
BeiT &  & \textbf{0.21} & \textbf{1.89} & \textbf{0.76} && 0.22 & 1.87 & 0.66\\
DeiT &  & 0.13 & 1.42 & 0.61 && 0.24 & 2.12 & 0.80\\
Swin &  & 0.14 & 1.47 & 0.63 && 0.21 & 1.85 & 0.65\\
xRegNet & & 0.14 & 1.50 & 0.64 && \textbf{0.33} & \textbf{2.54} & \textbf{0.88}\\
CLIP &  & 0.12 & 1.39 & 0.62 && 0.22 & 1.99 & 0.80\\
\midrule
ViT & \multirow{6}{5em}{leopard \\ cat} & 0.17 & 1.76 & 0.75 & \multirow{6}{5em}{angora \\ cat} & 0.31 & 2.39 & 0.83\\
BeiT & & \textbf{0.31} & \textbf{2.47} & \textbf{0.93} && 0.30 & 2.22 & 0.73\\
DeiT & & 0.15 & 1.47 & 0.60 && \textbf{0.32} & 2.48 & \textbf{0.86}\\
Swin & & 0.13 & 1.26 & 0.50 && \textbf{0.32} & \textbf{2.49} & \textbf{0.86}\\
xRegNet & & 0.18 & 1.70 & 0.69 && 0.28 & 2.22 & 0.78\\
CLIP & & 0.22 & 1.87 & 0.74 && 0.31 & 2.44 & \textbf{0.86}\\
\midrule
ViT & \multirow{6}{5em}{egyptian \\ cat} & 0.11 & 1.33 & 0.50 & \multirow{6}{5em}{cat} & 0.09 & 1.15 & 0.50\\
BeiT & & \textbf{0.16} & 1.63 & 0.60 && \textbf{0.23} & \textbf{1.81} & \textbf{0.68}\\
DeiT & & 0.11 & 1.31 & 0.50 && 0.05 & 0.72 & 0.29\\
Swin & & 0.11 & 1.34 & 0.52 && 0.05 & 0.73 & 0.29\\
xRegNet & & 0.10 & 1.24 & 0.46 && 0.06 & 0.89 & 0.39\\
CLIP & & 0.15 & \textbf{1.65} & \textbf{0.70} && 0.12 & 1.42 & 0.66\\
\bottomrule
\end{tabular}
\end{table}

The aforementioned claim regarding the need for \textit{fine-grained evaluation} is supported by demonstrating results using \textit{knowledge-driven metrics} based on conceptual distance as provided by WordNet (Tables \ref{tab:kdm}\& \ref{tab:kdm2}). Since higher path similarity/LCH, WUPS scores are better, we denote with bold best (higher) scores for each category.

By comparing path similarity, LCH, and WUPS metrics across categories, we observe that categories having a large number of irrelevant FP (marked in red in Tables \ref{tab:miscl}, \ref{tab:miscl2}), such as  'cougar cat' and 'lynx cat', followed by 'egyptian cat' and 'cat' also present low knowledge-driven metric scores in Tables \ref{tab:kdm}, \ref{tab:kdm2}, as expected. Other categories such as 'angora cat', 'leopard cat', and 'tiger cat' that present misclassifications of related (sibling or parent) categories also present higher knowledge-driven metric scores. Therefore, we can safely assume that employing knowledge-driven metrics for evaluating fine-grained classification results is highly correlated with human-interpretable notions of similarity and therefore trustworthy.

Model performance is rather clear when examining CNN classifiers. EfficientNet achieves predicting more relevant FP images compared to other classifiers for the majority of the categories. On the other hand, it is harder to draw a similar conclusion for Transformer-based classifiers, as different models perform better for different categories; however, compared to CNN classifiers the results of knowledge-driven metrics are the same or higher for most categories. Even though this difference is not impressive, transformer-based models showcase an improved capability of predicting more relevant classes, when failing to return the GT one. 

\section{Conclusion}
In this work, we implemented a novel distribution shift involving uncurated web images, upon which we tested convolutional and transformer-based image classifiers. Selecting closely related categories for classification is instructed by hierarchical knowledge, which is again employed to evaluate the quality of results. We prove that accuracy-related metrics can only scratch the surface of classification evaluation since they cannot capture semantic relationships between misclassified samples and ground truth labels. To this end, we propose an explainable, knowledge-driven evaluation scheme, able to quantify misclassification relevance by providing the semantic distance between false positive and real labels. The same scheme is also used to compare the classification capabilities of CNN vs transformer-based models on the implemented distribution shift. As future work, we plan to extend our analysis to more query terms in order to examine the extend of our current findings, and also combine the uncurated image classification setting with artificial corruptions to enhance our insights.

\begin{acknowledgments}
The research work was supported by the Hellenic Foundation for Research and Innovation (HFRI) under the 3rd Call for HFRI PhD Fellowships (Fellowship Number 5537).
\end{acknowledgments}

\bibliography{sample-ceur}

\begin{thebibliography}{53}
\expandafter\ifx\csname natexlab\endcsname\relax\def\natexlab#1{#1}\fi
\providecommand{\url}[1]{\texttt{#1}}
\providecommand{\href}[2]{#2}
\providecommand{\path}[1]{#1}
\providecommand{\DOIprefix}{doi:}
\providecommand{\ArXivprefix}{arXiv:}
\providecommand{\URLprefix}{URL: }
\providecommand{\Pubmedprefix}{pmid:}
\providecommand{\doi}[1]{\href{http://dx.doi.org/#1}{\path{#1}}}
\providecommand{\Pubmed}[1]{\href{pmid:#1}{\path{#1}}}
\providecommand{\bibinfo}[2]{#2}
\ifx\xfnm\relax \def\xfnm[#1]{\unskip,\space#1}\fi
\bibitem[{Deng et~al.(2009)Deng, Dong, Socher, Li, Li, and Fei-Fei}]{imagenet}
\bibinfo{author}{J.~Deng}, \bibinfo{author}{W.~Dong},
  \bibinfo{author}{R.~Socher}, \bibinfo{author}{L.-J. Li},
  \bibinfo{author}{K.~Li}, \bibinfo{author}{L.~Fei-Fei},
\newblock \bibinfo{title}{Imagenet: A large-scale hierarchical image database},
\newblock in: \bibinfo{booktitle}{2009 IEEE Conference on Computer Vision and
  Pattern Recognition}, \bibinfo{year}{2009}, pp. \bibinfo{pages}{248--255}.
  \DOIprefix\doi{10.1109/CVPR.2009.5206848}.
\bibitem[{Krizhevsky et~al.(2012)Krizhevsky, Sutskever, and Hinton}]{alexnet}
\bibinfo{author}{A.~Krizhevsky}, \bibinfo{author}{I.~Sutskever},
  \bibinfo{author}{G.~E. Hinton},
\newblock \bibinfo{title}{Imagenet classification with deep convolutional
  neural networks},
\newblock in: \bibinfo{booktitle}{Proceedings of the 25th International
  Conference on Neural Information Processing Systems - Volume 1}, NIPS'12,
  \bibinfo{publisher}{Curran Associates Inc.}, \bibinfo{address}{Red Hook, NY,
  USA}, \bibinfo{year}{2012}, p. \bibinfo{pages}{1097–1105}.
\bibitem[{Li et~al.(2020)Li, Yang, Peng, and Liu}]{image-classification-survey}
\bibinfo{author}{Z.~Li}, \bibinfo{author}{W.~Yang}, \bibinfo{author}{S.~Peng},
  \bibinfo{author}{F.~Liu},
\newblock \bibinfo{title}{A survey of convolutional neural networks: Analysis,
  applications, and prospects},
\newblock \bibinfo{publisher}{arXiv}, \bibinfo{year}{2020}. \URLprefix
  \url{https://arxiv.org/abs/2004.02806}.
  \DOIprefix\doi{10.48550/ARXIV.2004.02806}.
\bibitem[{Vaswani et~al.(2017)Vaswani, Shazeer, Parmar, Uszkoreit, Jones,
  Gomez, Kaiser, and Polosukhin}]{transformer}
\bibinfo{author}{A.~Vaswani}, \bibinfo{author}{N.~Shazeer},
  \bibinfo{author}{N.~Parmar}, \bibinfo{author}{J.~Uszkoreit},
  \bibinfo{author}{L.~Jones}, \bibinfo{author}{A.~N. Gomez},
  \bibinfo{author}{L.~Kaiser}, \bibinfo{author}{I.~Polosukhin},
  \bibinfo{title}{Attention is all you need}, \bibinfo{year}{2017}. \URLprefix
  \url{https://arxiv.org/abs/1706.03762}.
  \DOIprefix\doi{10.48550/ARXIV.1706.03762}.
\bibitem[{Dosovitskiy et~al.(2020)Dosovitskiy, Beyer, Kolesnikov, Weissenborn,
  Zhai, Unterthiner, Dehghani, Minderer, Heigold, Gelly, Uszkoreit, and
  Houlsby}]{vit}
\bibinfo{author}{A.~Dosovitskiy}, \bibinfo{author}{L.~Beyer},
  \bibinfo{author}{A.~Kolesnikov}, \bibinfo{author}{D.~Weissenborn},
  \bibinfo{author}{X.~Zhai}, \bibinfo{author}{T.~Unterthiner},
  \bibinfo{author}{M.~Dehghani}, \bibinfo{author}{M.~Minderer},
  \bibinfo{author}{G.~Heigold}, \bibinfo{author}{S.~Gelly},
  \bibinfo{author}{J.~Uszkoreit}, \bibinfo{author}{N.~Houlsby},
  \bibinfo{title}{An image is worth 16x16 words: Transformers for image
  recognition at scale}, \bibinfo{year}{2020}. \URLprefix
  \url{https://arxiv.org/abs/2010.11929}.
  \DOIprefix\doi{10.48550/ARXIV.2010.11929}.
\bibitem[{Yu et~al.(2022)Yu, Wang, Vasudevan, Yeung, Seyedhosseini, and
  Wu}]{coca}
\bibinfo{author}{J.~Yu}, \bibinfo{author}{Z.~Wang},
  \bibinfo{author}{V.~Vasudevan}, \bibinfo{author}{L.~Yeung},
  \bibinfo{author}{M.~Seyedhosseini}, \bibinfo{author}{Y.~Wu},
  \bibinfo{title}{Coca: Contrastive captioners are image-text foundation
  models}, \bibinfo{year}{2022}. \URLprefix
  \url{https://arxiv.org/abs/2205.01917}.
  \DOIprefix\doi{10.48550/ARXIV.2205.01917}.
\bibitem[{Wortsman et~al.(2022)Wortsman, Ilharco, Gadre, Roelofs,
  Gontijo-Lopes, Morcos, Namkoong, Farhadi, Carmon, Kornblith, and
  Schmidt}]{modelsoups}
\bibinfo{author}{M.~Wortsman}, \bibinfo{author}{G.~Ilharco},
  \bibinfo{author}{S.~Y. Gadre}, \bibinfo{author}{R.~Roelofs},
  \bibinfo{author}{R.~Gontijo-Lopes}, \bibinfo{author}{A.~S. Morcos},
  \bibinfo{author}{H.~Namkoong}, \bibinfo{author}{A.~Farhadi},
  \bibinfo{author}{Y.~Carmon}, \bibinfo{author}{S.~Kornblith},
  \bibinfo{author}{L.~Schmidt},
\newblock \bibinfo{title}{Model soups: averaging weights of multiple fine-tuned
  models improves accuracy without increasing inference time},
\newblock in: \bibinfo{editor}{K.~Chaudhuri}, \bibinfo{editor}{S.~Jegelka},
  \bibinfo{editor}{L.~Song}, \bibinfo{editor}{C.~Szepesvari},
  \bibinfo{editor}{G.~Niu}, \bibinfo{editor}{S.~Sabato} (Eds.),
  \bibinfo{booktitle}{Proceedings of the 39th International Conference on
  Machine Learning}, volume \bibinfo{volume}{162} of
  \textit{\bibinfo{series}{Proceedings of Machine Learning Research}},
  \bibinfo{publisher}{PMLR}, \bibinfo{year}{2022}, pp.
  \bibinfo{pages}{23965--23998}. \URLprefix
  \url{https://proceedings.mlr.press/v162/wortsman22a.html}.
\bibitem[{Liu et~al.(2021)Liu, Hu, Lin, Yao, Xie, Wei, Ning, Cao, Zhang, Dong,
  Wei, and Guo}]{swin2}
\bibinfo{author}{Z.~Liu}, \bibinfo{author}{H.~Hu}, \bibinfo{author}{Y.~Lin},
  \bibinfo{author}{Z.~Yao}, \bibinfo{author}{Z.~Xie}, \bibinfo{author}{Y.~Wei},
  \bibinfo{author}{J.~Ning}, \bibinfo{author}{Y.~Cao},
  \bibinfo{author}{Z.~Zhang}, \bibinfo{author}{L.~Dong},
  \bibinfo{author}{F.~Wei}, \bibinfo{author}{B.~Guo},
\newblock \bibinfo{title}{Swin transformer v2: Scaling up capacity and
  resolution}  (\bibinfo{year}{2021}). \URLprefix
  \url{https://arxiv.org/abs/2111.09883}.
  \DOIprefix\doi{10.48550/ARXIV.2111.09883}.
\bibitem[{Bao et~al.(2021)Bao, Dong, Piao, and Wei}]{beit}
\bibinfo{author}{H.~Bao}, \bibinfo{author}{L.~Dong}, \bibinfo{author}{S.~Piao},
  \bibinfo{author}{F.~Wei}, \bibinfo{title}{Beit: Bert pre-training of image
  transformers}, \bibinfo{year}{2021}. \URLprefix
  \url{https://arxiv.org/abs/2106.08254}.
  \DOIprefix\doi{10.48550/ARXIV.2106.08254}.
\bibitem[{Cohen et~al.(2019)Cohen, Rosenfeld, and Kolter}]{shift1}
\bibinfo{author}{J.~Cohen}, \bibinfo{author}{E.~Rosenfeld},
  \bibinfo{author}{Z.~Kolter},
\newblock \bibinfo{title}{Certified adversarial robustness via randomized
  smoothing},
\newblock in: \bibinfo{editor}{K.~Chaudhuri},
  \bibinfo{editor}{R.~Salakhutdinov} (Eds.), \bibinfo{booktitle}{Proceedings of
  the 36th International Conference on Machine Learning},
  volume~\bibinfo{volume}{97} of \textit{\bibinfo{series}{Proceedings of
  Machine Learning Research}}, \bibinfo{publisher}{PMLR}, \bibinfo{year}{2019},
  pp. \bibinfo{pages}{1310--1320}. \URLprefix
  \url{https://proceedings.mlr.press/v97/cohen19c.html}.
\bibitem[{Geirhos et~al.(2018)Geirhos, Rubisch, Michaelis, Bethge, Wichmann,
  and Brendel}]{robustness5}
\bibinfo{author}{R.~Geirhos}, \bibinfo{author}{P.~Rubisch},
  \bibinfo{author}{C.~Michaelis}, \bibinfo{author}{M.~Bethge},
  \bibinfo{author}{F.~A. Wichmann}, \bibinfo{author}{W.~Brendel},
  \bibinfo{title}{Imagenet-trained cnns are biased towards texture; increasing
  shape bias improves accuracy and robustness}, \bibinfo{year}{2018}.
  \URLprefix \url{https://arxiv.org/abs/1811.12231}.
  \DOIprefix\doi{10.48550/ARXIV.1811.12231}.
\bibitem[{Yang et~al.(2019)Yang, Wang, and Heinze-Deml}]{shift2}
\bibinfo{author}{F.~Yang}, \bibinfo{author}{Z.~Wang},
  \bibinfo{author}{C.~Heinze-Deml},
\newblock \bibinfo{title}{Invariance-inducing regularization using worst-case
  transformations suffices to boost accuracy and spatial robustness},
\newblock in: \bibinfo{editor}{H.~Wallach}, \bibinfo{editor}{H.~Larochelle},
  \bibinfo{editor}{A.~Beygelzimer}, \bibinfo{editor}{F.~d\textquotesingle
  Alch\'{e}-Buc}, \bibinfo{editor}{E.~Fox}, \bibinfo{editor}{R.~Garnett}
  (Eds.), \bibinfo{booktitle}{Advances in Neural Information Processing
  Systems}, volume~\bibinfo{volume}{32}, \bibinfo{publisher}{Curran Associates,
  Inc.}, \bibinfo{year}{2019}. \URLprefix
  \url{https://proceedings.neurips.cc/paper/2019/file/1d01bd2e16f57892f0954902899f0692-Paper.pdf}.
\bibitem[{Zhai et~al.(2019)Zhai, Puigcerver, Kolesnikov, Ruyssen, Riquelme,
  Lucic, Djolonga, Pinto, Neumann, Dosovitskiy, Beyer, Bachem, Tschannen,
  Michalski, Bousquet, Gelly, and Houlsby}]{shift3}
\bibinfo{author}{X.~Zhai}, \bibinfo{author}{J.~Puigcerver},
  \bibinfo{author}{A.~Kolesnikov}, \bibinfo{author}{P.~Ruyssen},
  \bibinfo{author}{C.~Riquelme}, \bibinfo{author}{M.~Lucic},
  \bibinfo{author}{J.~Djolonga}, \bibinfo{author}{A.~S. Pinto},
  \bibinfo{author}{M.~Neumann}, \bibinfo{author}{A.~Dosovitskiy},
  \bibinfo{author}{L.~Beyer}, \bibinfo{author}{O.~Bachem},
  \bibinfo{author}{M.~Tschannen}, \bibinfo{author}{M.~Michalski},
  \bibinfo{author}{O.~Bousquet}, \bibinfo{author}{S.~Gelly},
  \bibinfo{author}{N.~Houlsby}, \bibinfo{title}{A large-scale study of
  representation learning with the visual task adaptation benchmark},
  \bibinfo{year}{2019}. \URLprefix \url{https://arxiv.org/abs/1910.04867}.
  \DOIprefix\doi{10.48550/ARXIV.1910.04867}.
\bibitem[{Hendrycks and Dietterich(2019)}]{robustness4}
\bibinfo{author}{D.~Hendrycks}, \bibinfo{author}{T.~Dietterich},
  \bibinfo{title}{Benchmarking neural network robustness to common corruptions
  and perturbations}, \bibinfo{year}{2019}.
  \href{http://arxiv.org/abs/1903.12261}{{\tt arXiv:1903.12261}}.
\bibitem[{Taori et~al.(2020)Taori, Dave, Shankar, Carlini, Recht, and
  Schmidt}]{robustness1}
\bibinfo{author}{R.~Taori}, \bibinfo{author}{A.~Dave},
  \bibinfo{author}{V.~Shankar}, \bibinfo{author}{N.~Carlini},
  \bibinfo{author}{B.~Recht}, \bibinfo{author}{L.~Schmidt},
  \bibinfo{title}{Measuring robustness to natural distribution shifts in image
  classification}, \bibinfo{year}{2020}. \URLprefix
  \url{https://arxiv.org/abs/2007.00644}.
  \DOIprefix\doi{10.48550/ARXIV.2007.00644}.
\bibitem[{Hendrycks et~al.(2020)Hendrycks, Basart, Mu, Kadavath, Wang, Dorundo,
  Desai, Zhu, Parajuli, Guo, Song, Steinhardt, and Gilmer}]{robustness3}
\bibinfo{author}{D.~Hendrycks}, \bibinfo{author}{S.~Basart},
  \bibinfo{author}{N.~Mu}, \bibinfo{author}{S.~Kadavath},
  \bibinfo{author}{F.~Wang}, \bibinfo{author}{E.~Dorundo},
  \bibinfo{author}{R.~Desai}, \bibinfo{author}{T.~L. Zhu},
  \bibinfo{author}{S.~Parajuli}, \bibinfo{author}{M.~Guo},
  \bibinfo{author}{D.~X. Song}, \bibinfo{author}{J.~Steinhardt},
  \bibinfo{author}{J.~Gilmer},
\newblock \bibinfo{title}{The many faces of robustness: A critical analysis of
  out-of-distribution generalization},
\newblock \bibinfo{journal}{2021 IEEE/CVF International Conference on Computer
  Vision (ICCV)}  (\bibinfo{year}{2020}) \bibinfo{pages}{8320--8329}.
\bibitem[{Barbu et~al.(2019)Barbu, Mayo, Alverio, Luo, Wang, Gutfreund,
  Tenenbaum, and Katz}]{robustness7}
\bibinfo{author}{A.~Barbu}, \bibinfo{author}{D.~Mayo},
  \bibinfo{author}{J.~Alverio}, \bibinfo{author}{W.~Luo},
  \bibinfo{author}{C.~Wang}, \bibinfo{author}{D.~Gutfreund},
  \bibinfo{author}{J.~Tenenbaum}, \bibinfo{author}{B.~Katz},
\newblock \bibinfo{title}{Objectnet: A large-scale bias-controlled dataset for
  pushing the limits of object recognition models},
\newblock in: \bibinfo{editor}{H.~Wallach}, \bibinfo{editor}{H.~Larochelle},
  \bibinfo{editor}{A.~Beygelzimer}, \bibinfo{editor}{F.~d\textquotesingle
  Alch\'{e}-Buc}, \bibinfo{editor}{E.~Fox}, \bibinfo{editor}{R.~Garnett}
  (Eds.), \bibinfo{booktitle}{Advances in Neural Information Processing
  Systems}, volume~\bibinfo{volume}{32}, \bibinfo{publisher}{Curran Associates,
  Inc.}, \bibinfo{year}{2019}. \URLprefix
  \url{https://proceedings.neurips.cc/paper/2019/file/97af07a14cacba681feacf3012730892-Paper.pdf}.
\bibitem[{Hendrycks et~al.(2019)Hendrycks, Zhao, Basart, Steinhardt, and
  Song}]{robustness8}
\bibinfo{author}{D.~Hendrycks}, \bibinfo{author}{K.~Zhao},
  \bibinfo{author}{S.~Basart}, \bibinfo{author}{J.~Steinhardt},
  \bibinfo{author}{D.~Song}, \bibinfo{title}{Natural adversarial examples},
  \bibinfo{year}{2019}. \URLprefix \url{https://arxiv.org/abs/1907.07174}.
  \DOIprefix\doi{10.48550/ARXIV.1907.07174}.
\bibitem[{DeVries and Taylor(2017)}]{train0}
\bibinfo{author}{T.~DeVries}, \bibinfo{author}{G.~W. Taylor},
  \bibinfo{title}{Improved regularization of convolutional neural networks with
  cutout}, \bibinfo{year}{2017}. \URLprefix
  \url{https://arxiv.org/abs/1708.04552}.
  \DOIprefix\doi{10.48550/ARXIV.1708.04552}.
\bibitem[{Zheng et~al.(2016)Zheng, Song, Leung, and Goodfellow}]{train1}
\bibinfo{author}{S.~Zheng}, \bibinfo{author}{Y.~Song},
  \bibinfo{author}{T.~Leung}, \bibinfo{author}{I.~Goodfellow},
  \bibinfo{title}{Improving the robustness of deep neural networks via
  stability training}, \bibinfo{year}{2016}. \URLprefix
  \url{https://arxiv.org/abs/1604.04326}.
  \DOIprefix\doi{10.48550/ARXIV.1604.04326}.
\bibitem[{Taylor and Nitschke(2017)}]{train4}
\bibinfo{author}{L.~Taylor}, \bibinfo{author}{G.~Nitschke},
  \bibinfo{title}{Improving deep learning using generic data augmentation},
  \bibinfo{year}{2017}. \URLprefix \url{https://arxiv.org/abs/1708.06020}.
  \DOIprefix\doi{10.48550/ARXIV.1708.06020}.
\bibitem[{Rebuffi et~al.(2021)Rebuffi, Gowal, Calian, Stimberg, Wiles, and
  Mann}]{train2}
\bibinfo{author}{S.-A. Rebuffi}, \bibinfo{author}{S.~Gowal},
  \bibinfo{author}{D.~A. Calian}, \bibinfo{author}{F.~Stimberg},
  \bibinfo{author}{O.~Wiles}, \bibinfo{author}{T.~Mann}, \bibinfo{title}{Data
  augmentation can improve robustness}, \bibinfo{year}{2021}. \URLprefix
  \url{https://arxiv.org/abs/2111.05328}.
  \DOIprefix\doi{10.48550/ARXIV.2111.05328}.
\bibitem[{Fellbaum(1998)}]{wordnet}
\bibinfo{author}{C.~Fellbaum},
\newblock \bibinfo{title}{Wordnet: An electronic lexical database}
  (\bibinfo{year}{1998}).
\bibitem[{Simonyan and Zisserman(2014)}]{vgg}
\bibinfo{author}{K.~Simonyan}, \bibinfo{author}{A.~Zisserman},
\newblock \bibinfo{title}{Very deep convolutional networks for large-scale
  image recognition},
\newblock \bibinfo{journal}{CoRR} \bibinfo{volume}{abs/1409.1556}
  (\bibinfo{year}{2014}).
\bibitem[{Szegedy et~al.(2016)Szegedy, Vanhoucke, Ioffe, Shlens, and
  Wojna}]{inception}
\bibinfo{author}{C.~Szegedy}, \bibinfo{author}{V.~Vanhoucke},
  \bibinfo{author}{S.~Ioffe}, \bibinfo{author}{J.~Shlens},
  \bibinfo{author}{Z.~Wojna},
\newblock \bibinfo{title}{Rethinking the inception architecture for computer
  vision},
\newblock in: \bibinfo{booktitle}{2016 IEEE Conference on Computer Vision and
  Pattern Recognition (CVPR)}, \bibinfo{year}{2016}, pp.
  \bibinfo{pages}{2818--2826}. \DOIprefix\doi{10.1109/CVPR.2016.308}.
\bibitem[{He et~al.(2016)He, Zhang, Ren, and Sun}]{resnet}
\bibinfo{author}{K.~He}, \bibinfo{author}{X.~Zhang}, \bibinfo{author}{S.~Ren},
  \bibinfo{author}{J.~Sun},
\newblock \bibinfo{title}{Deep residual learning for image recognition},
\newblock in: \bibinfo{booktitle}{2016 IEEE Conference on Computer Vision and
  Pattern Recognition (CVPR)}, \bibinfo{year}{2016}, pp.
  \bibinfo{pages}{770--778}. \DOIprefix\doi{10.1109/CVPR.2016.90}.
\bibitem[{Chollet(2016)}]{Xception}
\bibinfo{author}{F.~Chollet},
\newblock \bibinfo{title}{Xception: Deep learning with depthwise separable
  convolutions},
\newblock \bibinfo{journal}{2017 IEEE Conference on Computer Vision and Pattern
  Recognition (CVPR)}  (\bibinfo{year}{2016}) \bibinfo{pages}{1800--1807}.
\bibitem[{Szegedy et~al.(2017)Szegedy, Ioffe, Vanhoucke, and
  Alemi}]{inceptionresnet}
\bibinfo{author}{C.~Szegedy}, \bibinfo{author}{S.~Ioffe},
  \bibinfo{author}{V.~Vanhoucke}, \bibinfo{author}{A.~A. Alemi},
\newblock \bibinfo{title}{Inception-v4, inception-resnet and the impact of
  residual connections on learning},
\newblock in: \bibinfo{booktitle}{Proceedings of the Thirty-First AAAI
  Conference on Artificial Intelligence}, AAAI'17, \bibinfo{publisher}{AAAI
  Press}, \bibinfo{year}{2017}, p. \bibinfo{pages}{4278–4284}.
\bibitem[{Liu et~al.(2022)Liu, Mao, Wu, Feichtenhofer, Darrell, and
  Xie}]{convnext}
\bibinfo{author}{Z.~Liu}, \bibinfo{author}{H.~Mao}, \bibinfo{author}{C.~Wu},
  \bibinfo{author}{C.~Feichtenhofer}, \bibinfo{author}{T.~Darrell},
  \bibinfo{author}{S.~Xie},
\newblock \bibinfo{title}{A convnet for the 2020s},
\newblock \bibinfo{journal}{CoRR} \bibinfo{volume}{abs/2201.03545}
  (\bibinfo{year}{2022}). \URLprefix \url{https://arxiv.org/abs/2201.03545}.
  \href{http://arxiv.org/abs/2201.03545}{{\tt arXiv:2201.03545}}.
\bibitem[{Khan et~al.(2022)Khan, Naseer, Hayat, Zamir, Khan, and
  Shah}]{vit-survey}
\bibinfo{author}{S.~Khan}, \bibinfo{author}{M.~Naseer},
  \bibinfo{author}{M.~Hayat}, \bibinfo{author}{S.~W. Zamir},
  \bibinfo{author}{F.~S. Khan}, \bibinfo{author}{M.~Shah},
\newblock \bibinfo{title}{Transformers in vision: A survey},
\newblock \bibinfo{journal}{{ACM} Computing Surveys} \bibinfo{volume}{54}
  (\bibinfo{year}{2022}) \bibinfo{pages}{1--41}. \URLprefix
  \url{https://doi.org/10.1145\%2F3505244}. \DOIprefix\doi{10.1145/3505244}.
\bibitem[{Touvron et~al.(2021)Touvron, Cord, Douze, Massa, Sablayrolles, and
  Jegou}]{deit}
\bibinfo{author}{H.~Touvron}, \bibinfo{author}{M.~Cord},
  \bibinfo{author}{M.~Douze}, \bibinfo{author}{F.~Massa},
  \bibinfo{author}{A.~Sablayrolles}, \bibinfo{author}{H.~Jegou},
\newblock \bibinfo{title}{Training data-efficient image transformers \&
  distillation through attention},
\newblock in: \bibinfo{editor}{M.~Meila}, \bibinfo{editor}{T.~Zhang} (Eds.),
  \bibinfo{booktitle}{Proceedings of the 38th International Conference on
  Machine Learning}, volume \bibinfo{volume}{139} of
  \textit{\bibinfo{series}{Proceedings of Machine Learning Research}},
  \bibinfo{publisher}{PMLR}, \bibinfo{year}{2021}, pp.
  \bibinfo{pages}{10347--10357}.
\bibitem[{Liu et~al.(2022)Liu, Hu, Lin, Yao, Xie, Wei, Ning, Cao, Zhang, Dong,
  Wei, and Guo}]{swin}
\bibinfo{author}{Z.~Liu}, \bibinfo{author}{H.~Hu}, \bibinfo{author}{Y.~Lin},
  \bibinfo{author}{Z.~Yao}, \bibinfo{author}{Z.~Xie}, \bibinfo{author}{Y.~Wei},
  \bibinfo{author}{J.~Ning}, \bibinfo{author}{Y.~Cao},
  \bibinfo{author}{Z.~Zhang}, \bibinfo{author}{L.~Dong},
  \bibinfo{author}{F.~Wei}, \bibinfo{author}{B.~Guo},
\newblock \bibinfo{title}{Swin transformer v2: Scaling up capacity and
  resolution},
\newblock in: \bibinfo{booktitle}{2022 IEEE/CVF Conference on Computer Vision
  and Pattern Recognition (CVPR)}, \bibinfo{year}{2022}, pp.
  \bibinfo{pages}{11999--12009}. \DOIprefix\doi{10.1109/CVPR52688.2022.01170}.
\bibitem[{Radosavovic et~al.(2020)Radosavovic, Kosaraju, Girshick, He, and
  Dollár}]{regnet}
\bibinfo{author}{I.~Radosavovic}, \bibinfo{author}{R.~P. Kosaraju},
  \bibinfo{author}{R.~Girshick}, \bibinfo{author}{K.~He},
  \bibinfo{author}{P.~Dollár}, \bibinfo{title}{Designing network design
  spaces}, \bibinfo{year}{2020}.
\bibitem[{Paul and Chen(2021)}]{robustvit}
\bibinfo{author}{S.~Paul}, \bibinfo{author}{P.-Y. Chen},
\newblock \bibinfo{title}{Vision transformers are robust learners},
\newblock in: \bibinfo{booktitle}{AAAI Conference on Artificial Intelligence},
  \bibinfo{year}{2021}.
\bibitem[{Recht et~al.(2019)Recht, Roelofs, Schmidt, and Shankar}]{robustness0}
\bibinfo{author}{B.~Recht}, \bibinfo{author}{R.~Roelofs},
  \bibinfo{author}{L.~Schmidt}, \bibinfo{author}{V.~Shankar},
\newblock \bibinfo{title}{Do imagenet classifiers generalize to imagenet?},
\newblock in: \bibinfo{booktitle}{International Conference on Machine
  Learning}, \bibinfo{year}{2019}.
\bibitem[{Geirhos et~al.(2018)Geirhos, Temme, Rauber, Schütt, Bethge, and
  Wichmann}]{robustness6}
\bibinfo{author}{R.~Geirhos}, \bibinfo{author}{C.~R.~M. Temme},
  \bibinfo{author}{J.~Rauber}, \bibinfo{author}{H.~H. Schütt},
  \bibinfo{author}{M.~Bethge}, \bibinfo{author}{F.~A. Wichmann},
  \bibinfo{title}{Generalisation in humans and deep neural networks},
  \bibinfo{year}{2018}. \URLprefix \url{https://arxiv.org/abs/1808.08750}.
  \DOIprefix\doi{10.48550/ARXIV.1808.08750}.
\bibitem[{Laugros et~al.(2021)Laugros, Caplier, and Ospici}]{robustness2}
\bibinfo{author}{A.~Laugros}, \bibinfo{author}{A.~Caplier},
  \bibinfo{author}{M.~Ospici},
\newblock \bibinfo{title}{Using synthetic corruptions to measure robustness to
  natural distribution shifts},
\newblock \bibinfo{journal}{ArXiv} \bibinfo{volume}{abs/2107.12052}
  (\bibinfo{year}{2021}).
\bibitem[{Hendrycks et~al.(2019)Hendrycks, Zhao, Basart, Steinhardt, and
  Song}]{nae}
\bibinfo{author}{D.~Hendrycks}, \bibinfo{author}{K.~Zhao},
  \bibinfo{author}{S.~Basart}, \bibinfo{author}{J.~Steinhardt},
  \bibinfo{author}{D.~Song}, \bibinfo{title}{Natural adversarial examples},
  \bibinfo{year}{2019}. \URLprefix \url{https://arxiv.org/abs/1907.07174}.
  \DOIprefix\doi{10.48550/ARXIV.1907.07174}.
\bibitem[{Biggio et~al.(2013)Biggio, Corona, Maiorca, Nelson, {\v{S}
  }rndi{\'{c}}, Laskov, Giacinto, and Roli}]{adversarial3}
\bibinfo{author}{B.~Biggio}, \bibinfo{author}{I.~Corona},
  \bibinfo{author}{D.~Maiorca}, \bibinfo{author}{B.~Nelson},
  \bibinfo{author}{N.~{\v{S} }rndi{\'{c}}}, \bibinfo{author}{P.~Laskov},
  \bibinfo{author}{G.~Giacinto}, \bibinfo{author}{F.~Roli},
\newblock \bibinfo{title}{Evasion attacks against machine learning at test
  time},
\newblock in: \bibinfo{booktitle}{Advanced Information Systems Engineering},
  \bibinfo{publisher}{Springer Berlin Heidelberg}, \bibinfo{year}{2013}, pp.
  \bibinfo{pages}{387--402}. \URLprefix
  \url{https://doi.org/10.1007%2F978-3-642-40994-3_25}.
  \DOIprefix\doi{10.1007/978-3-642-40994-3_25}.
\bibitem[{Dong et~al.(2020)Dong, Fu, Yang, Pang, Su, Xiao, and
  Zhu}]{adversarial1}
\bibinfo{author}{Y.~Dong}, \bibinfo{author}{Q.-A. Fu},
  \bibinfo{author}{X.~Yang}, \bibinfo{author}{T.~Pang},
  \bibinfo{author}{H.~Su}, \bibinfo{author}{Z.~Xiao}, \bibinfo{author}{J.~Zhu},
\newblock \bibinfo{title}{Benchmarking adversarial robustness on image
  classification},
\newblock in: \bibinfo{booktitle}{2020 IEEE/CVF Conference on Computer Vision
  and Pattern Recognition (CVPR)}, \bibinfo{year}{2020}, pp.
  \bibinfo{pages}{318--328}. \DOIprefix\doi{10.1109/CVPR42600.2020.00040}.
\bibitem[{Tsipras et~al.(2018)Tsipras, Santurkar, Engstrom, Turner, and
  Madry}]{adversarial5}
\bibinfo{author}{D.~Tsipras}, \bibinfo{author}{S.~Santurkar},
  \bibinfo{author}{L.~Engstrom}, \bibinfo{author}{A.~Turner},
  \bibinfo{author}{A.~Madry}, \bibinfo{title}{Robustness may be at odds with
  accuracy}, \bibinfo{year}{2018}. \URLprefix
  \url{https://arxiv.org/abs/1805.12152}.
  \DOIprefix\doi{10.48550/ARXIV.1805.12152}.
\bibitem[{Ozbulak et~al.(2021)Ozbulak, Anzaku, Neve, and Messem}]{adversarial2}
\bibinfo{author}{U.~Ozbulak}, \bibinfo{author}{E.~T. Anzaku},
  \bibinfo{author}{W.~D. Neve}, \bibinfo{author}{A.~V. Messem},
\newblock \bibinfo{title}{Selection of source images heavily influences the
  effectiveness of adversarial attacks},
\newblock \bibinfo{journal}{ArXiv} \bibinfo{volume}{abs/2106.07141}
  (\bibinfo{year}{2021}).
\bibitem[{Wang et~al.(2022)Wang, Ullah, Mianjy, and Arora}]{adversarial4}
\bibinfo{author}{Y.~Wang}, \bibinfo{author}{E.~Ullah},
  \bibinfo{author}{P.~Mianjy}, \bibinfo{author}{R.~Arora},
\newblock \bibinfo{title}{Adversarial robustness is at odds with lazy
  training},
\newblock \bibinfo{journal}{ArXiv} \bibinfo{volume}{abs/2207.00411}
  (\bibinfo{year}{2022}).
\bibitem[{Pinto et~al.(2022)Pinto, Torr, and Dokania}]{impartial}
\bibinfo{author}{F.~Pinto}, \bibinfo{author}{P.~H.~S. Torr},
  \bibinfo{author}{P.~K. Dokania},
\newblock \bibinfo{title}{An impartial take to the cnn vs transformer
  robustness contest},
\newblock in: \bibinfo{booktitle}{European Conference on Computer Vision},
  \bibinfo{year}{2022}.
\bibitem[{Wang et~al.(2022)Wang, Bai, Zhou, and Xie}]{cnn-or-transformer}
\bibinfo{author}{Z.~Wang}, \bibinfo{author}{Y.~Bai}, \bibinfo{author}{Y.~Zhou},
  \bibinfo{author}{C.~Xie},
\newblock \bibinfo{title}{Can cnns be more robust than transformers?},
\newblock \bibinfo{journal}{ArXiv} \bibinfo{volume}{abs/2206.03452}
  (\bibinfo{year}{2022}).
\bibitem[{Bhojanapalli et~al.(2021)Bhojanapalli, Chakrabarti, Glasner, Li,
  Unterthiner, and Veit}]{understand-robust}
\bibinfo{author}{S.~Bhojanapalli}, \bibinfo{author}{A.~Chakrabarti},
  \bibinfo{author}{D.~Glasner}, \bibinfo{author}{D.~Li},
  \bibinfo{author}{T.~Unterthiner}, \bibinfo{author}{A.~Veit},
  \bibinfo{title}{Understanding robustness of transformers for image
  classification}, \bibinfo{year}{2021}.
\bibitem[{Deng et~al.(2022)Deng, Gould, and Zheng}]{understand-robust2}
\bibinfo{author}{W.~Deng}, \bibinfo{author}{S.~Gould},
  \bibinfo{author}{L.~Zheng},
\newblock \bibinfo{title}{On the strong correlation between model invariance
  and generalization},
\newblock \bibinfo{journal}{ArXiv} \bibinfo{volume}{abs/2207.07065}
  (\bibinfo{year}{2022}).
\bibitem[{Mintun et~al.(2021)Mintun, Kirillov, and Xie}]{understand-robust3}
\bibinfo{author}{E.~Mintun}, \bibinfo{author}{A.~Kirillov},
  \bibinfo{author}{S.~Xie},
\newblock \bibinfo{title}{On interaction between augmentations and corruptions
  in natural corruption robustness},
\newblock in: \bibinfo{editor}{M.~Ranzato}, \bibinfo{editor}{A.~Beygelzimer},
  \bibinfo{editor}{Y.~Dauphin}, \bibinfo{editor}{P.~Liang},
  \bibinfo{editor}{J.~W. Vaughan} (Eds.), \bibinfo{booktitle}{Advances in
  Neural Information Processing Systems}, volume~\bibinfo{volume}{34},
  \bibinfo{publisher}{Curran Associates, Inc.}, \bibinfo{year}{2021}, pp.
  \bibinfo{pages}{3571--3583}. \URLprefix
  \url{https://proceedings.neurips.cc/paper/2021/file/1d49780520898fe37f0cd6b41c5311bf-Paper.pdf}.
\bibitem[{Sandler et~al.(2018)Sandler, Howard, Zhu, Zhmoginov, and
  Chen}]{mobilenet}
\bibinfo{author}{M.~Sandler}, \bibinfo{author}{A.~Howard},
  \bibinfo{author}{M.~Zhu}, \bibinfo{author}{A.~Zhmoginov},
  \bibinfo{author}{L.-C. Chen},
\newblock \bibinfo{title}{Mobilenetv2: Inverted residuals and linear
  bottlenecks},
\newblock \bibinfo{year}{2018}, pp. \bibinfo{pages}{4510--4520}.
  \DOIprefix\doi{10.1109/CVPR.2018.00474}.
\bibitem[{Zoph et~al.(2018)Zoph, Vasudevan, Shlens, and Le}]{nasnet}
\bibinfo{author}{B.~Zoph}, \bibinfo{author}{V.~Vasudevan},
  \bibinfo{author}{J.~Shlens}, \bibinfo{author}{Q.~Le},
\newblock \bibinfo{title}{Learning transferable architectures for scalable
  image recognition},
\newblock \bibinfo{year}{2018}, pp. \bibinfo{pages}{8697--8710}.
  \DOIprefix\doi{10.1109/CVPR.2018.00907}.
\bibitem[{Huang et~al.(2017)Huang, Liu, van~der Maaten, and
  Weinberger}]{densenet}
\bibinfo{author}{G.~Huang}, \bibinfo{author}{Z.~Liu},
  \bibinfo{author}{L.~van~der Maaten}, \bibinfo{author}{K.~Weinberger},
\newblock \bibinfo{title}{Densely connected convolutional networks},
\newblock \bibinfo{year}{2017}. \DOIprefix\doi{10.1109/CVPR.2017.243}.
\bibitem[{Tan and Le(2019)}]{efficientnet}
\bibinfo{author}{M.~Tan}, \bibinfo{author}{Q.~Le},
\newblock \bibinfo{title}{Efficientnet: Rethinking model scaling for
  convolutional neural networks},
\newblock \bibinfo{year}{2019}.
\bibitem[{Radford et~al.(2021)Radford, Kim, Hallacy, Ramesh, Goh, Agarwal,
  Sastry, Askell, Mishkin, Clark, Krueger, and Sutskever}]{clip}
\bibinfo{author}{A.~Radford}, \bibinfo{author}{J.~W. Kim},
  \bibinfo{author}{C.~Hallacy}, \bibinfo{author}{A.~Ramesh},
  \bibinfo{author}{G.~Goh}, \bibinfo{author}{S.~Agarwal},
  \bibinfo{author}{G.~Sastry}, \bibinfo{author}{A.~Askell},
  \bibinfo{author}{P.~Mishkin}, \bibinfo{author}{J.~Clark},
  \bibinfo{author}{G.~Krueger}, \bibinfo{author}{I.~Sutskever},
\newblock \bibinfo{title}{Learning transferable visual models from natural
  language supervision},
\newblock in: \bibinfo{booktitle}{International Conference on Machine
  Learning}, \bibinfo{year}{2021}.

\end{thebibliography}

\appendix
\section{More CNN misclassifications}
In Table \ref{tab:miscl-cnn2}, we present the continuation of the results present in Table \ref{tab:miscl-cnn} for the rest of the CNN models presenting non-zero accuracy. It becomes evident that the capacity of the classifier plays an important role in identifying relevant FP: MobileNetV2, which already demonstrated low accuracy scores, also fail to retrieve semantically related FP classes. This can be easily observed from the numerous red entries corresponding to this model.

Other than that, the results agree with the observations analyzed in Table \ref{tab:miscl-cnn}, where 'egyptian cat' label demonstrated many irrelevant FP, contrary to 'tabby cat' or 'tiger cat' labels.

\begin{table}[htp!]
\caption{(Continuation of Tab. \ref{tab:miscl-cnn}).
Common misclassifications for selected GT cat classes and misclassification frequency for CNNs.}
\label{tab:miscl-cnn2}
\begin{tabular}{>{\centering\arraybackslash}p{2.3em}|>
{\centering\arraybackslash}p{3em}|>{\centering\arraybackslash}p{7.5em}>{\centering\arraybackslash}p{2.3em}|>{\centering\arraybackslash}p{7.5em}>{\centering\arraybackslash}p{2.3em}|>{\centering\arraybackslash}p{5.9em}>{\centering\arraybackslash}p{2em}}
\toprule
&  & \multicolumn{2}{c}{Top-1} &  \multicolumn{2}{c}{Top-2} &  \multicolumn{2}{c}{Top-3} \\
\cline{3-8}
Model & GT & FP & MF & FP & MF & FP & MF \\
\toprule
\multirow{10}{3em}{VGG\\19} & tabby & \textcolor{blue}{egyptian cat} & 28.00\% &
\textcolor{blue}{tiger cat} & 20.00\%& lynx & 16.00\%\\
& angora & \textcolor{blue}{persian cat} & 34.78\% & \textcolor{red}{arctic fox} & 10.87\% & \textcolor{blue}{egyptian cat} & 10.87\%\\
&lynx & egyptian cat & 20.00\% & \textcolor{red}{coyote} & 20.00\% & \textcolor{red}{timber wolf} & 20.00\%\\
& siamese & \textcolor{red}{whippet } & 16.67\% & \textcolor{red}{fur coat} & 16.67\% & \textcolor{blue}{egyptian cat} & 16.67\%\\
& tiger& \textcolor{blue}{tabby cat} & 34.88\% & \textcolor{blue}{egyptian cat} & 16.28\% & tiger & 13.95\%\\
& persian & lynx & 20.00\% & \textcolor{red}{pekinese} & 25.00\% & \textcolor{red}{fur coat} & 10.00\%\\
& cougar & \textcolor{blue}{lynx} & 45.45\% & \textcolor{red}{coyote} & 18.18\% & \textcolor{red}{timber wolf} & 9.09\%\\
&leopard & egyptian cat & 46.00\% & \textcolor{blue}{lynx} & 16.00\% &
 jaguar & 12.00\%\\
& egyptian & lynx & 8.11\% & \textcolor{red}{mask} & 5.41\% & \textcolor{red}{book jacket} & 5.41\%\\
& cat & \textcolor{red}{fur coat} & 11.11\% & snow leopard & 11.11\% & \textcolor{red}{mousetrap} & 11.11\%\\
\midrule
\multirow{10}{6em}{Mobile\\Net\\V2} & tabby & \textcolor{red}{comic book} & 14.29\% & \textcolor{red}{mask} & 10.20\% & \textcolor{red}{sock} & 8.16\%\\
& angora & \textcolor{red}{shower curtain} & 20.00\% & \textcolor{red}{window screen} & 16.00 & \textcolor{red}{spotlight} & 8.00\% \\
& lynx & \textcolor{red}{west highland white terrier}  & 8.00\% & tiger & 6.00\% & \textcolor{red}{traffic light} & 6.00\% \\
& siamese & \textcolor{red}{shower curtain} & 14.58\% & \textcolor{red}{sock} & 14.58\% & \textcolor{red}{mask} & 14.58\%\\
& tiger & \textcolor{red}{zebra} & 11.86\% & \textcolor{red}{mask} & 6.78\% & \textcolor{red}{maze} & 6.78\% \\
& persian & \textcolor{red}{spotlight} & 10.87\% & \textcolor{red}{shower curtain} & 8.70\% & \textcolor{red}{ant} & 8.70\% \\
& cougar & \textcolor{red}{comic book} & 14.00\% & \textcolor{red}{mask} & 8.00\%& \textcolor{red}{theater curtain } & 8.00\% \\
& leopard & \textcolor{red}{knot} & 14.00\% & tiger & 12.00\% & \textcolor{red}{mask}  & 6.00\% \\
& egyptian & \textcolor{red}{windsor tie } & 10.00\% & \textcolor{red}{theater curtain} & 10.00\% & \textcolor{red}{spotlight}  & 8.00\% \\
& cat & \textcolor{red}{shower curtain} & 10.53\% & \textcolor{red}{window screen} & 7.89\% & \textcolor{red}{teddy} & 7.89\% \\
\midrule
\multirow{12}{5em}{Effic\\ient\\Net} & tabby & \textcolor{blue}{tiger cat} & 62.22\% & \textcolor{blue}{egyptian cat} & 28.89\% & \textcolor{blue}{persian cat} & 4.44\% \\
& angora & \textcolor{blue}{persian cat} & 72.92\% & \textcolor{blue}{egyptian cat} & 20.83\% & \textcolor{blue}{tabby cat} & 2.08\% \\
& lynx & egyptian cat & 60.00\% & \textcolor{blue}{tiger cat} & 20.00\% & tabby cat & 20.00\\
& siamese & \textcolor{blue}{egyptian cat} & 100.0\% & - & - & - & - \\
& tiger & \textcolor{blue}{egyptian cat} & 34.21\% & \textcolor{blue}{tabby cat} & 18.42\% & tiger & 15.79\% \\
& persian & \textcolor{blue}{tabby cat} & 100.0\%  & - & - & - & -\\
& cougar & \textcolor{blue}{tiger cat} & 100.0\% & - & - & - & -\\
& leopard & egyptian cat & 56.00\% & \textcolor{blue}{lynx} & 22.00\% & \textcolor{blue}{tiger cat} & 16.00\% \\
& egyptian & \textcolor{red}{comic book } & 13.33\% & \textcolor{red}{mexican hairless} & 13.33\% & \textcolor{red}{lampshade } & 6.67\% \\
& cat & \textcolor{red}{macaque} & 33.33\% & \textcolor{red}{mexican hairless} & 33.33\% & \textcolor{red}{indigo bunting } & 33.33\%\\
\midrule
\multirow{10}{5em}{Conv\\Next} & tabby & \textcolor{blue}{tiger cat} & 75.00\% & \textcolor{blue}{egyptian cat} & 15.00\% & \textcolor{red}{web site} & 5.00\%\\
& angora & \textcolor{blue}{persian cat} & 46.67\% & \textcolor{blue}{egyptian cat} & 35.56  & \textcolor{blue}{tabby cat} & 6.67\% \\
& lynx & tabby cat & 75.00 & \textcolor{blue}{tiger cat} & 25.00\% & - & - \\
& siamese & \textcolor{blue}{egyptian cat} & 75.00\%  & \textcolor{red}{golden retriever} & 25.00\% & - & -\\
& tiger & \textcolor{blue}{tabby cat} & 31.82\% & \textcolor{blue}{egyptian cat} & 31.82\% & tiger & 9.09\% \\
& persian & \textcolor{blue}{siamese cat} & 100.0\% & -& - & - & -\\
& cougar & \textcolor{red}{web site} & 100.0\% & -& -& - & -\\
& leopard & egyptian cat & 32.00\% & \textcolor{blue}{lynx} & 18.00\% & leopard & 16.00\% \\
& egyptian  & \textcolor{red}{mexican hairless} & 20.83\% &  \textcolor{red}{mask} & 12.50\% & \textcolor{red}{comic book } & 12.50\%\\
& cat & \textcolor{red}{fur coat} & 50.00\% & \textcolor{red}{mexican hairless} & 50.00\% & - & -\\
\bottomrule
\end{tabular}
\end{table}

\end{document}